%% file: 0main.tex
\documentclass[journal, comsoc, 10pt]{IEEEtran}
\IEEEoverridecommandlockouts
\usepackage{amsmath,amssymb,amsfonts,mathrsfs}
\usepackage{algorithmic}
\usepackage{graphicx}
\usepackage{textcomp}
\usepackage{color}
\usepackage{balance}
\usepackage{enumitem}
\usepackage[normalem]{ulem}
\usepackage{xspace}
\usepackage{multirow}

\newtheorem{definition}{Definition}[section]
\newtheorem{property}{Property}[section]

\makeatletter
\newcommand{\linebreakand}{%
  \end{@IEEEauthorhalign}
  \hfill\mbox{}\par
  \mbox{}\hfill\begin{@IEEEauthorhalign}
}

\newcommand{\pni}{$90^{\text{th}}$\xspace}
\newcommand{\pnf}{$95^{\text{th}}$\xspace}
\newcommand{\pnn}{$99^{\text{th}}$\xspace}

\makeatother
\def\BibTeX{{\rm B\kern-.05em{\sc i\kern-.025em b}\kern-.08em
    T\kern-.1667em\lower.7ex\hbox{E}\kern-.125emX}}

\begin{document}
\bstctlcite{IEEEexample:BSTcontrol}

\title{SafeTail: Efficient Tail Latency Optimization in Edge Service Scheduling via Computational Redundancy Management}

\author{Jyoti Shokhanda, Utkarsh Pal, Aman Kumar, Soumi Chattopadhyay, Arani Bhattacharya
\thanks{\textit{Jyoti Shokhanda, Utkarsh Pal, Aman Kumar and Arani Bhattacharya are affiliated to Indraprastha Institute of Information Technology Delhi, New Delhi, email: \{jyotis, utkarsh20144, aman20279, arani\}@iiitd.ac.in.\\
Soumi Chattopadhyay is with the Department of Computer Science and Engineering, Indian Institute of Technology Indore, India, e-mail: soumi@iiti.ac.in.
This work has been submitted to the IEEE for possible publication. Copyright may be transferred without notice, after which this version may no longer be accessible.
}
}
}
\newcommand{\notesc}[1]{\textbf{\textcolor{red}{SC: #1}}}
\maketitle

\begin{abstract}
Optimizing tail latency while efficiently managing computational resources is crucial for delivering high-performance, latency-sensitive services in edge computing. Emerging applications, such as augmented reality, require low-latency computing services with high reliability on user devices, which often have limited computational capabilities. Consequently, these devices depend on nearby edge servers for processing. However, inherent uncertainties in network and computation latencies—stemming from variability in wireless networks and fluctuating server loads—make service delivery on time challenging.
Existing approaches often focus on optimizing median latency but fall short of addressing the specific challenges of tail latency in edge environments, particularly under uncertain network and computational conditions. Although some methods do address tail latency, they typically rely on fixed or excessive redundancy and lack adaptability to dynamic network conditions, often being designed for cloud environments rather than the unique demands of edge computing.
In this paper, we introduce SafeTail, a framework that meets both median and tail response time targets, with tail latency defined as latency beyond the \pni percentile threshold. SafeTail addresses this challenge by selectively replicating services across multiple edge servers to meet target latencies. SafeTail employs a reward-based deep learning framework to learn optimal placement strategies, balancing the need to achieve target latencies with minimizing additional resource usage. Through trace-driven simulations, SafeTail demonstrated near-optimal performance and outperformed most baseline strategies across three diverse services.

\end{abstract}


\input{1intro}
\input{2Motivation}

\input{3Framework}

\input{4Implementation}
\input{5Results}
\input{6relatedWork}
\input{7conclusion}

\bibliographystyle{IEEEtran}

\bibliography{reference}
\end{document}

%% file: 1intro.tex
\section{Introduction}
\noindent
In the realm of edge computing \cite{jiang2021mobile}, latency-sensitive applications play a crucial role in providing seamless and high-quality user experiences. Technologies such as augmented reality (AR) \cite{siriwardhana2021survey}, virtual reality (VR) \cite{wang2018service}, and real-time video conferencing demand exceptionally low latency to ensure responsiveness and fluid interaction \cite{zhang2019improving}. For example, AR applications used in interactive gaming or navigation require near-instantaneous processing to align digital overlays with the real world, while VR experiences depend on minimal latency to create immersive, lag-free environments. Real-time video conferencing tools also require rapid data transmission to maintain clear and uninterrupted communication. These applications often run on user devices with limited computation power, relying on nearby edge servers for efficient processing.

Conversely, some latency-sensitive applications can tolerate higher median latency but still require stringent control over tail latency. For instance, in batch processing systems for data analytics, large-scale data analysis or scientific simulations may process data in batches and provide results at periodic intervals rather than in real-time \cite{shekhar2018performance}. While the system can accept higher median latency for routine tasks, it is crucial to manage tail latency carefully, especially for urgent queries or emergency analyses. Ensuring that tail latency remains within acceptable limits is vital for maintaining the system's effectiveness and responsiveness during critical instances, as excessive delays in these scenarios could significantly impact the application's performance and reliability. Thus, while overall throughput and accuracy are paramount, controlling tail latency is essential to meet the performance requirements of latency-sensitive applications.

A major difficulty in supporting latency-sensitive applications is ensuring the timely delivery of services with an acceptable stable median latency and minimal deviation from that latency.
However, consistently achieving acceptable latency levels remains challenging. Service requests involve both network and computation latencies, each with inherent uncertainties that make it difficult to meet target latencies reliably.
Consequently, most existing research focuses on optimizing median or mean latencies with edge servers but often overlooks higher percentiles of latency, such as the \pni, \pnf, and \pnn percentiles. These higher percentiles are crucial for delivering a high-quality user experience in latency-sensitive applications \cite{tail-eurosys,Tail_Guard}. Many studies on edge service scheduling fail to address these tail latencies effectively. The issue of tail latency is particularly pronounced in edge computing \cite{latency-reliability}, where both the computation on edge servers and communication over wireless networks are subject to significant variability. Therefore, addressing high tail latency through the effective use of edge servers is essential for maintaining service quality in latency-sensitive applications.


Prior research has explored task scheduling on edge servers using deep reinforcement learning (DRL) \cite{subrat-paper}. These studies demonstrate that DRL can reduce task completion times by rewarding schedules with lower latency. However, their focus has primarily been on standard tasks offloaded from smartphones, rather than latency-sensitive tasks. Latency-sensitive applications impose strict constraints not only on median or mean latency but also on its entire distribution. Inadequate handling of these constraints can result in significant safety issues \cite{safety_V2V}. Consequently, these studies do not address the impact of DRL on tail latency.



One of the primary techniques for reducing tail latency involves introducing redundancy. For instance, a user device may submit a service to multiple edge servers, allowing it to utilize the fastest response among them. While this redundancy improves tail latencies, it can also increase the utilization of edge computing resources, such as network bandwidth and costs. Therefore, it is crucial to manage redundancy carefully to minimize tail latency while controlling resource usage.


However, determining where to place services with controlled redundancy remains complex. As redundancy increases, the number of potential scheduling options grows exponentially. For a single-end device with $n$ available edge servers, the number of possible schedules is $2^{n} - 1$, making exhaustive search impractical for service request execution.
 

We tackle this challenge by first identifying a target latency for each service, and then designing our reward to minimize the actual difference from the target. SafeTail, our framework, uses redundant scheduling combined with a reward-based deep-learning approach to address tail latency. This method involves assigning the same service request to multiple edge servers and enabling the system to learn from experience and interaction, considering both computation and network latencies. Unlike methods that rely on fixed redundancy or focus solely on optimizing one type of latency, SafeTail dynamically adjusts redundancy based on real-time conditions. This strategy reduces overall latency by effectively managing both network and computation latency, allowing the user device to select the fastest response and thus improving overall system responsiveness.



Our experiments begin with an in-depth tail latency analysis of the YOLOv5 object detection service on our own WiFi network. We observed that network latency is influenced by factors like the number of active users, while computation latency is affected by compute resource availability, memory, and server configuration. These variables introduce uncertainty in service latency~\cite{joint_service}, further compounded by varying workloads on edge servers and network unpredictability~\cite{dynamic-workload}.


{Our experiments depend on the collection of our own WiFi network and compute traces. 
We then assessed the performance of our reward-based deep learning mechanism on the median as well as tail latency values for three distinct services: 
(i) Object detection using YOLOv5, 
(ii) Instance segmentation of images, and 
(iii) Removal of noise from audio.
These three services are all latency-sensitive, and some of them are often used in real-time applications such as virtual reality and video conferencing.
We evaluated SafeTail using trace-driven simulations on a system with five edge servers and compared its performance against four baseline techniques. Our findings show that SafeTail effectively reduces both median and tail latency compared to these baseline methods. 
We now summarize our contributions as follows:



{\emph{(i)}} We address the challenge of optimizing service latency while managing redundancy in edge computing, where uncertainties are higher than in cloud computing. Specifically, we account for uncertainties in both wireless networks—affecting network latency, including transmission and propagation latencies, network load, and computation latencies. While similar issues have been tackled in cloud environments, the inherent variability in edge networks requires a more specialized approach for latency-sensitive services. Our approach focuses on achieving optimal tail latency by prioritizing latency optimization over resource utilization, i.e., the number of edge servers selected for redundant service execution, while still incorporating controlled redundancy based on the dynamic states of edge servers.

{\emph{(ii)}} We propose a reward-based deep learning framework that optimizes latency through adaptively controlled redundancy. This approach automatically balances the need to meet target latencies with the efficient use of resources.

{\emph{(iii)}} We developed a real-world testbed and collected execution traces from three distinct applications: (a) single-shot object detection from images, (b) instance segmentation of images, and (c) noise removal from audio under varying network and edge load conditions. Using these traces for simulation, we demonstrate that our reward-based deep learning framework significantly optimizes both median and tail latency.
 
The rest of this paper is organized as follows: Section II presents the mathematical formulation of the problem, while Section III provides a motivating scenario for this research. In Section IV, we describe the proposed methodology. Section V outlines the experimental setup and details our analysis of the results. Section VI offers a review of relevant literature, and Section VII discusses the limitations and potential directions for future work. Finally, Section VIII concludes the paper.

%% file: 2Motivation.tex
\section{Problem Formulation}
\noindent
In this section, we formulate our problem mathematically.
Our framework has the following inputs:
\begin{itemize}[leftmargin=*]

\item A set of homogeneous edge servers ${\mathcal{E}} = \{ e_1, e_2, \ldots, e_n \}$.

\item The dynamic state of each edge server $e_i \in {\mathcal{E}}$ at any timestep $t$ is identified by 5-tuple: 
$(\lambda^{u(t)}_i, \lambda^{d(t)}_i, {\mathcal{M}}_i^t, {\mathcal{U}}_i^t, \ell^t_i)$, 
where $\lambda^{u(t)}_i, \lambda^{d(t)}_i, {\mathcal{M}}_i^t, {\mathcal{U}}_i^t, \ell_i^t$ 
denote the uplink and downlink bandwidths, memory and CPU utilization, and number of active users accessing $e_i$ at timestep $t$, respectively. In the rest of this paper, however, we omit the superscript $t$ from each symbol to represent the values of the above parameters as experienced by the user when the service is actually requested.

\item A user at a specific location with the requirement of edge servers for the execution of a service $s_j$.

\item A user is denoted by a 3-tuple: $U = (L, \Lambda^u, \Lambda^d)$, where $L$, $\Lambda^u, \Lambda^d$ represent the location, upload and download bandwidths of the user device, respectively.

\item Each service $s_j$ is characterized by 3-tuple: $({\mathcal{I}}_j, {\mathcal{O}}_j, \Gamma_j)$, 
where ${\mathcal{I}}_j$ represents a set of input parameters required by the service, ${\mathcal{O}}_j$ denotes a set of output parameters generated by the service after execution, and $\Gamma_j$ includes the characteristics of the input parameters of that influence the computation time of $s_j$. It may be noted that $\Gamma_j$ varies across different services. 
\end{itemize}

\noindent
The objective of our work is to reduce the overall latency for the service $s_j$. 
Before going to discuss the details of our framework, we first explain the key components of latency.
%
Multiple events are associated with the execution of a service on an edge server, which actually contributes to latency computation, as discussed below.

\begin{itemize}[leftmargin=*]
\item {\emph{Transfer of service-input}}: A set of input parameters ${\mathcal{I}}_j$ for the service $s_j$ must be transferred from the user device to the edge server for execution. This transfer time includes both transmission and propagation latency. The transfer process involves three key events: (i) uploading the input file from the user device, (ii) propagating the file from the user device to the edge server (represented by a function of edge server and user location $\rho_i (L)$), and (iii) downloading the input file to the edge server.
 
 \item {\emph{Execution of service}}: Once an edge server receives the input of a service, the service is executed on it. The computation time ${\mathcal{C}}(e_i, s_j)$ at $e_i$ is a function of various parameters of $s_j$ and the dynamic state of the edge server $e_i$. 
 
 \item {\emph{Transfer of service output}}: The output parameters ${\mathcal{O}}_j$ of the service $s_j$ are transferred from the edge server to the user device, similar to the process of transfer of the service input.
\end{itemize}

\noindent
Considering a service is executed in $e_i$, service latency is mathematically defined as follows. The symbols used in Eq. \eqref{eq:lat} are defined in brackets after introducing each term above.
\begin{gather}\scriptsize
{\cal{L}}_{i} = 
\underbrace{\left(\underbrace{\frac{Size(\mathcal{I}_j)}{\min(\lambda^d_i, \Lambda^u)}}_\text{{uploading}} + \underbrace{\rho_i(L)}_\text{propagation latency} + \underbrace{\frac{Size(\mathcal{I}_j)}{\min(\lambda^d_i, \Lambda^u)}}_\text{downloading}\right)}_\text{Transfer of service input} + \underbrace{\mathcal{C}(e_i, s_j)}_\text{Computation latency} + \nonumber \\
\scriptsize
\underbrace{\left(\underbrace{\frac{Size(\mathcal{O}_j)}{\min(\lambda^u_i, \Lambda^d)}}_\text{{uploading}} + \underbrace{\rho_i(L)}_\text{propagation latency} + \underbrace{\frac{Size(\mathcal{O}_j)}{\min(\lambda^u_i, \Lambda^d)}}_\text{downloading}\right)}_\text{Transfer of service output}
\label{eq:lat}
\end{gather}

\noindent

\begin{figure*}[!t]
 \centering
 \includegraphics[width=0.215\textwidth]{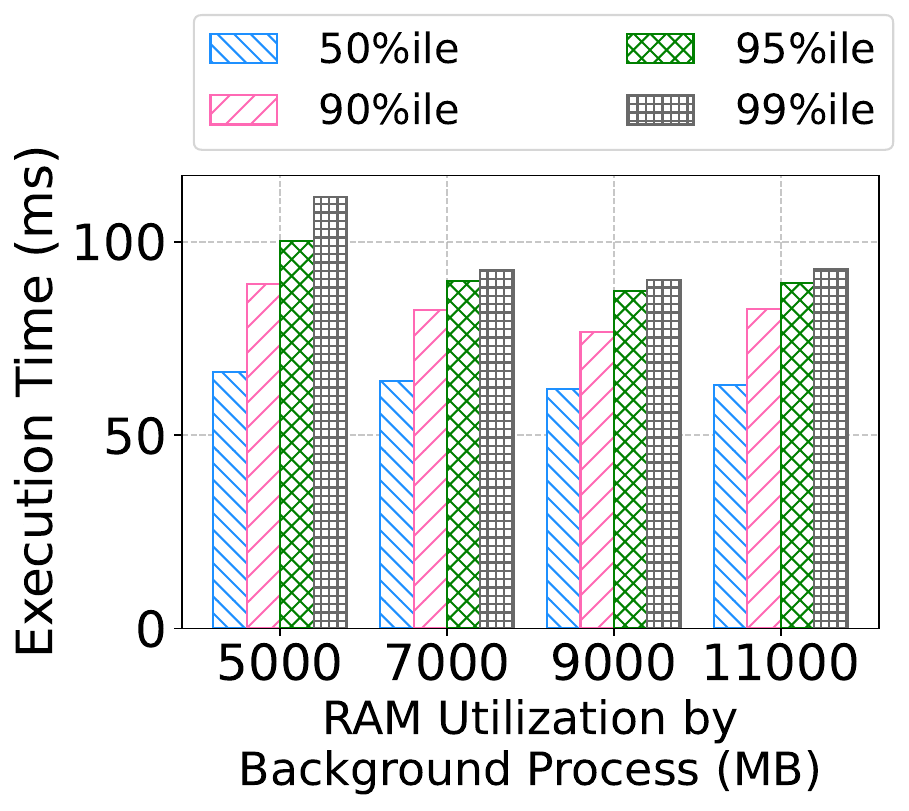}
 \includegraphics[width=0.23\textwidth]{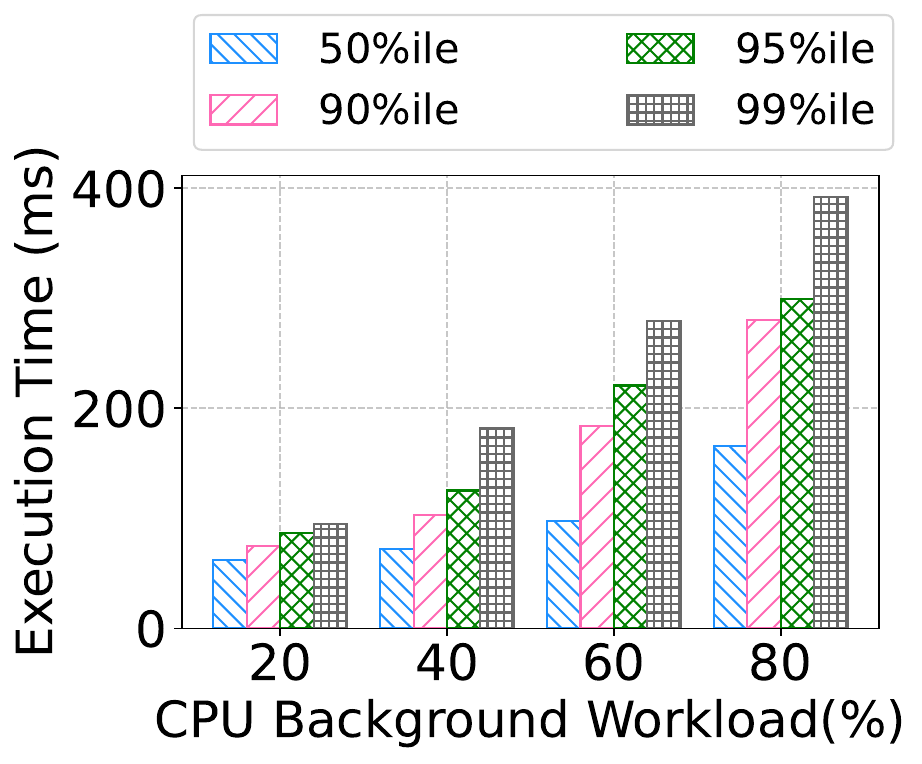}
 \includegraphics[width=0.23\textwidth]{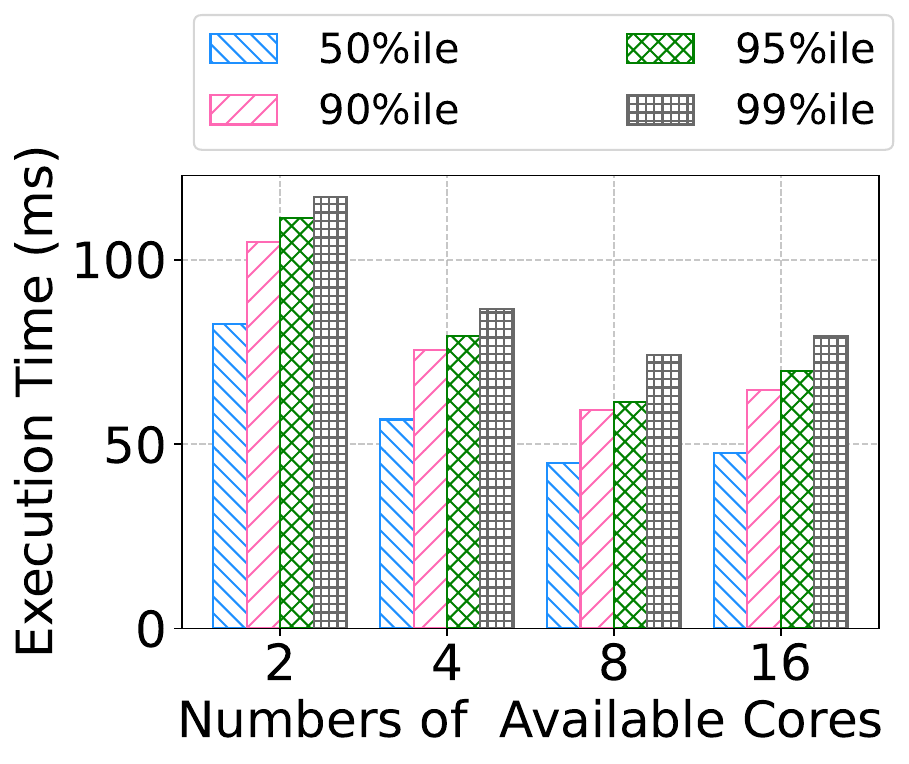}
 \includegraphics[width=0.23\textwidth]{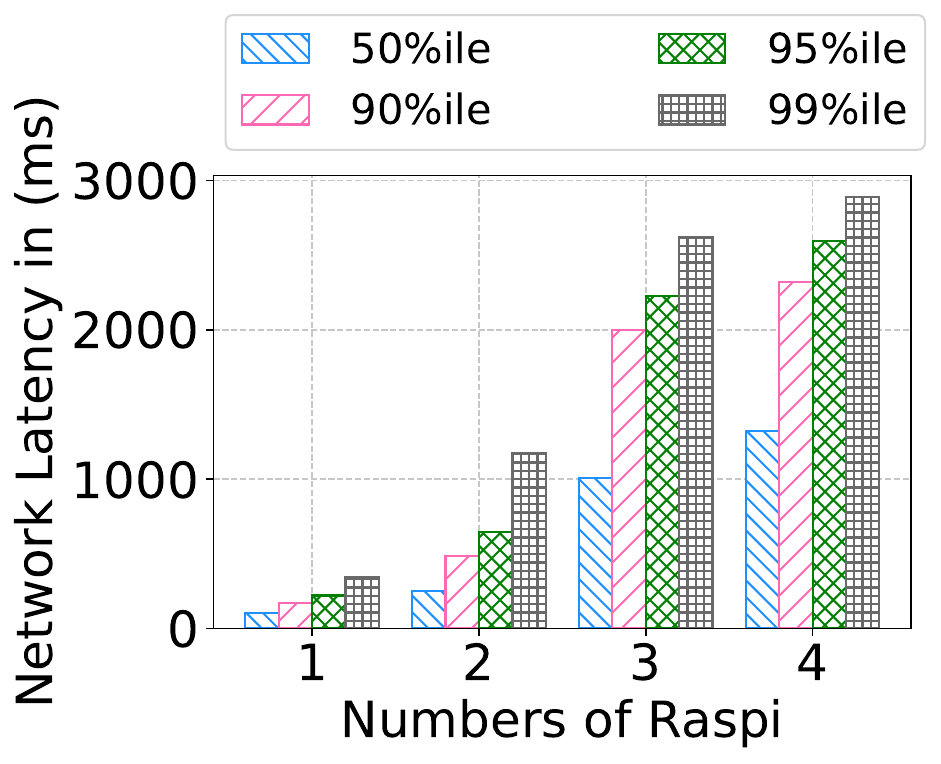}\\
 (a)~~~~~~~~~~~~~~~~~~~~~~~~~~~~~(b)~~~~~~~~~~~~~~~~~~~~~~~~~~~~~~~~~(c)~~~~~~~~~~~~~~~~~~~~~~~~~~~~~(d)
     \caption{
     Latency characteristics for the YOLOv5 object detection service: Illustration of variation in latency with changes in (a) RAM utilization by background processes, (b) CPU background workload, (c) the number of available cores, and (d) the number of Raspberry Pi devices to simulate varying network load.
     }
     \label{fig:mot}
 \end{figure*}

\noindent
In this paper, we have the following assumptions:

\emph{(i)} Each edge server can accept a certain number of requests. Beyond this limit, the edge server refuses to accept additional services.

\emph{(ii)} All resources in an edge server are equally distributed among the active users currently accessing it. In other words, we do not account for any waiting time in our current model.

\emph{(iii)} The number of edge servers ($n$) reachable from a user is reasonably small, which is most likely fewer than 10 in number. This is because the network used has a relatively smaller range, making it realistic to assume that only a small number of edge servers would be within this range.    

\emph{(iv)} A lightweight version of the service runs on the user device, with the latency tolerance of the service response bounded by the execution time of this lightweight version. However, due to its lower accuracy, the lightweight version is less preferred. Instead, we prioritize using the response obtained from the edge server, where the standard version of the service is executed.

In this paper, we aim to reduce long tail latency in edge computing to mitigate its severe impact by introducing controlled redundancy. 
In the next section, we explore the causes of high tail latency and its impact through an empirical study.

\section{Analyzing Tail Latency: Empirical Studies}
\noindent
In this section, we first characterize tail latency through experimental studies by observing it under different edge computing environments, including variations in compute and network loads.
Here, we select a widely used application in autonomous systems—object detection using YOLOv5—and run it on a desktop machine equipped with an Intel Core i7-11700 processor, 16GB of RAM, and a total of 16 virtual cores. We disabled the background processes for this experiment. We used the tools stress-ng \cite{stress-ng} to generate workload and task set \cite{tasket} to vary the number of cores allocated to the process from the total cores available on the system.
Fig. \ref{fig:mot} illustrates the different percentiles of latency and its various components (i.e., transmission latency, propagation latency, and computation latency) under varying network and compute parameters. Unless otherwise specified, we maintain the RAM at 16GB, the number of cores at 16, and the CPU background workload at 0\%.
We then varied a single parameter for each experiment, where the parameters are: (a) amount of available RAM, (b) CPU background workload, (c) number of available cores, and (d) number of devices communicating over the network. We repeated each experiment 1000 times and reported the results. We now discuss our empirical findings concerning tail latency.

\noindent \textbf{Effect of Changes in Available RAM:} In this experiment, we study the behavior of computation latency with respect to RAM availability. We varied the RAM utilization from 5000 to 11000 MB using stress-ng to generate workloads. Figure \ref{fig:mot}(a) shows the median and tail computation latencies (at the \pni, \pnf, and \pnn percentiles) for different levels of RAM utilization. As indicated in Fig. \ref{fig:mot}(a), the gap between median latency and tail latencies increased as RAM availability decreased. Moreover, higher percentile latencies were more significantly impacted by lower RAM availability compared to median latency.

\noindent \textbf{Effect of CPU Background Workload:} In this experiment (refer to Fig. \ref{fig:mot}(b)), we varied background workload on CPUs from 20\% to 80\%, and observed the changes in computation latency. We found a similar trend in median and tail latencies for this experiment as well.

\noindent \textbf{Effect of Number of Cores:} For this experiment, we varied the number of available cores from 2 to 16 and studied the behavior of computation latency (refer to Fig. \ref{fig:mot}(c)). We observed that computation latency decreased with an increase in the number of cores up to a certain threshold (here, up to 8 cores). Beyond this threshold, additional cores did not affect the computation latency.

\noindent \textbf{Effect of Network Load:} We generated network load by using additional Raspberry Pi nodes that sent data via iPerf 3 \cite{mortimer2018iperf3}. We then sent ping probes from our machine to a server over the same WiFi network to obtain our network latency. Figure \ref{fig:mot}(d) shows how network latencies change with varying computational resources (i.e., the number of Raspberry Pi nodes). We observed that latency increases only slightly with a small number of Raspberry Pi nodes (up to 2) but rises rapidly beyond this threshold. Other observations were consistent with the first and second experiments.

\noindent\textbf{The Challenge of Optimizing Tail Latency}: 
The above experiments demonstrate that tail latency is unevenly influenced by various resource parameters. For instance, parameters like the availability of CPU cores impact both median and tail latency similarly. In contrast, factors such as RAM availability, background workload, and network load have a more pronounced effect on tail latency compared to median latency. Therefore, we design a framework focused on minimizing tail latency, taking into account the varying sensitivities of these resource parameters.

To tackle this challenge, we employ several strategies. First, we use redundant service scheduling to mitigate latency unpredictability. Second, we develop a placement mechanism based on reward-based deep learning that learns from the dynamic state of edge servers to determine where to place the service. This approach ensures that latency-sensitive services are not reliant on simplistic scheduling models.

%% file: 3Framework.tex
\section{Framework and Methodology}
\noindent
In this section, we introduce SafeTail, our framework that utilizes a reward-based deep learning approach inspired by reinforcement learning. SafeTail is a user-centric service allocation framework that aims to reduce service latency by incorporating redundancy into the process. The primary motivation for our work is based on the premise that if one edge server fails to deliver a service response within the promised time, other edge servers may still be able to deliver the output on time. However, replicating the service execution to all edge servers can lead to increased network congestion and unnecessary resource utilization. 
Therefore, our objective in this paper is to minimize the service latency by determining the reasonable number of redundancies for a service execution across various servers, depending on the dynamic state of the edge servers, without significantly increasing resource utilization or compromising network traffic. We now discuss the details of our framework, starting with an introduction to the concept of redundancy scheduling for a service.

The goal of redundancy scheduling is to duplicate the execution of a service $s_j$ across multiple edge servers. By selecting a subset of edge servers ${\mathcal{E}}^{t}_k \subseteq {\mathcal{E}}$ at timestep $t$, we intend to minimize latency variability and achieve the fastest response. The rationale is that if one edge server is heavily loaded, another might have sufficient resources to execute the service within an acceptable time limit. This approach effectively addresses challenges such as uncertain propagation latencies, long transmission latencies, and variable computation latencies. The improved latency achieved by SafeTail through redundancy scheduling is demonstrated using Eq. \ref{eq:lat_redundancy}.
\begin{gather}\scriptsize
{\cal{L}}_{R} = \min_{e_i \in {\mathcal{E}}_k} \left(
{\cal{L}}_{i}\right)
\label{eq:lat_redundancy}
\end{gather}

\begin{figure}[!t]
\centering
\includegraphics[width=\linewidth]{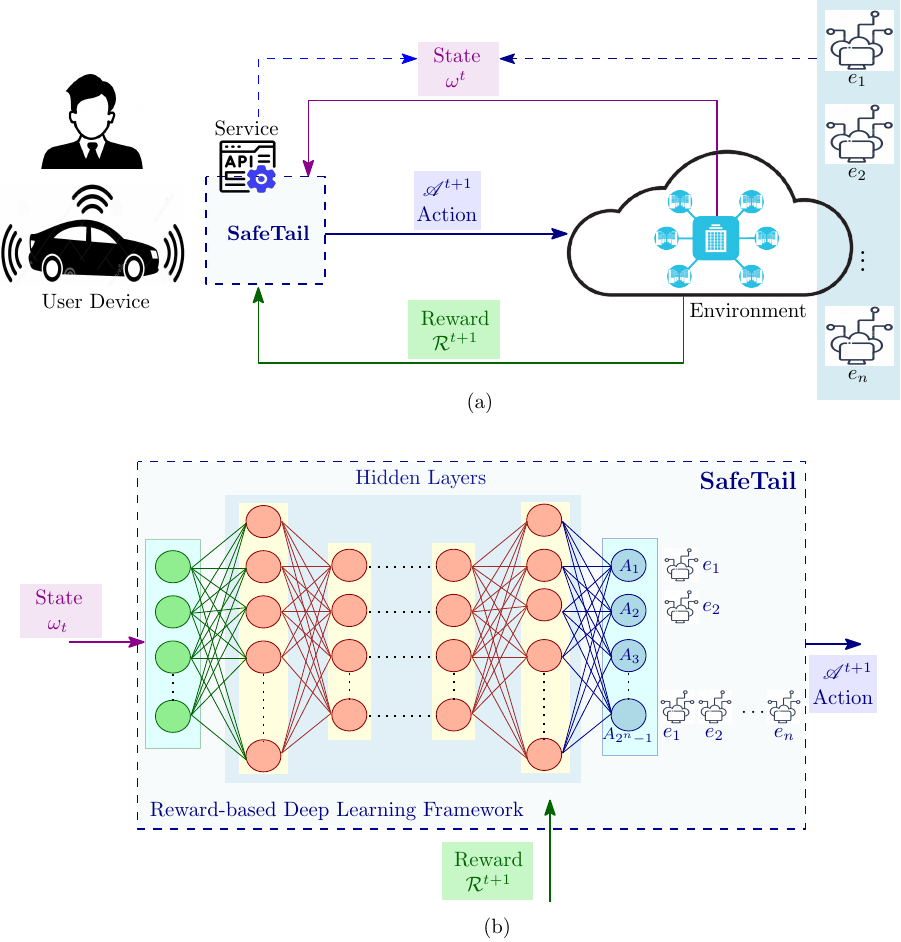}
    \caption{Overview of (a) SafeTail; (b) Reward-based Deep Learning Framework}
    \label{fig:SafeTail}
\end{figure}

\noindent
We now present our framework, SafeTail, which dynamically adapts redundant scheduling. SafeTail leverages a reward-based deep learning framework to approximate complex functions and understand the relationships among the dynamic state of the edge servers at timestep $t$, the service requirements, and the redundant executions that need to be introduced at timestep $t + 1$. The primary idea is to comprehend the dynamic state of each server and the service requirements to determine the expected latency for the service when executed on an edge server and the uncertainty of achieving that latency. Based on this, SafeTail decides the subset of edge servers needed to achieve the desired latency. Fig. \ref{fig:SafeTail} presents an overview of our framework. We discuss each component of SafeTail below. We begin by discussing the {\emph{state}} that depicts the dynamic condition of all edge servers, along with the components of the service on which SafeTail bases its {\emph{actions}}.

\begin{definition}{State}:
The state of the environment at timestep $t$, denoted by $\omega^t$, is defined by an $(n+1)$-tuple $(\Omega_1^t, \Omega_2^t, \ldots, \Omega_n^t, {\mathcal{P}}_j, U)$, where: 
$$\Omega_i^t = (\lambda^{u(t)}_i, \lambda^{d(t)}_i, {\mathcal{M}}_i^t, {\mathcal{U}}_i^t, \ell_i^t, \rho_i^\mu(L)); ~~\forall i \in \{1, 2, \ldots, n\}$$
$${\mathcal{P}}_j = (Size({\mathcal{I}}_j), eSize({\mathcal{O}}_j), \Gamma_j)$$
where the first five elements of $\Omega_i^t$ captures the dynamic state of edge server $e_i$ at timestep $t$. $\rho_i^\mu(L)$ represents the median propagation latency from the user device to $e_i$. ${\mathcal{P}}_j$ denotes the properties of the service $s_j$ that may influence the latency. Here, $Size({\mathcal{I}}_j)$ represents the size of ${\mathcal{I}}_j$ and $eSize({\mathcal{O}}_j)$ represents the estimated size of ${\mathcal{O}}_j$.
\hfill$\blacksquare$
\end{definition}

\noindent
It is important to note that various parameters in $\omega^t$ contribute to estimating different components of the latency. For instance, $\lambda^{u(t)}_i, \lambda^{d(t)}_i$ in $\Omega_i^t$ and $Size({\mathcal{I}}_j), eSize({\mathcal{O}}_j)$ in ${\mathcal{P}}_j$ helps estimate the transmission latency, while $\rho_i^\mu(L)$ in $\Omega_i^t$ enables to measure the propagation latency. Finally, ${\mathcal{M}}_i^t, {\mathcal{U}}_i^t, \ell_i^t$ in $\Omega_i^t$ and $\Gamma_j$ in ${\mathcal{P}}_j$ lead towards approximating the computation latency.

Once our learning network receives the state $\omega^t$ at $t$, it determines an appropriate {\emph{action}} at $(t+1)^{th}$ timestep, denoted by ${\mathscr{A}}^{t+1}$, by selecting a subset of edge servers for redundancy scheduling for $s_j$. For $n$ edge servers, there are $2^n-1$ possible actions, denoted as $A = \{A_1, A_2, \ldots, A_{2^n-1}\}$. Each action $A_i \in A$ is an element of ${\mathbb{P}}({\mathcal{E}}) \setminus \phi$, where ${\mathbb{P}}({\mathcal{E}})$ represents the powerset of ${\mathcal{E}}$. SafeTail employs an $\epsilon$-greedy exploration and exploitation strategy to choose an action.

\begin{equation}
 {\mathscr{A}}^{t+1} = 
 \begin{cases}
    \underset{A_i \in  A}{Random}(A_i)  & \text{ with probability } \epsilon\\
    \arg \underset{A_i \in  A}{\max} (A_i)  & \text{ with probability } (1-\epsilon)\\
 \end{cases} \label{eq:action}
\end{equation}

\noindent
where $\epsilon$ is a parameter that controls the balance between exploration and exploitation. Initially, $\epsilon$ is set to 1 when the network is untrained. SafeTail begins with 100\% exploration, where a random action is chosen. Each time SafeTail actuates an action, a reward is given based on the effectiveness of the action in optimizing tail latency and resource utilization. The state-action-reward tuple, i.e., $(\omega^t, {\mathscr{A}}^{t+1}, {\mathcal{R}}^{t+1})$ is stored to train the network. Once a sufficient number of tuples, denoted as $\kappa$, have been collected, we start training the network. 

Each time the network is partially trained, $\epsilon$ is decreased by a certain quantity, denoted as $\epsilon_{decay}$, until it reaches a minimum value, denoted as $\epsilon_{min}$. As $\epsilon$ is decreased, SafeTail continues its exploration and exploitation $\kappa$ times before retraining the network.
During exploitation, SafeTail invokes the reward-based learning network to return the action determined by the network, i.e., the second case of Eq. \eqref{eq:action}.  

Before discussing the architecture of the reward-based learning network, we first address the reward function of SafeTail. As previously mentioned, the reward is decided based on how effectively the action optimizes tail latency and resource utilization. Ideally, the highest reward is given when optimal latency is achieved with minimal resource usage. However, determining the optimal latency value without executing the service on all edge servers is infeasible. Therefore, we define a \emph{target latency}, which is a heuristic value that we approximate. In calculating the reward, we consider the achieved latency relative to this target latency. To proceed with defining the reward function, we first need to establish the target latency.

\begin{definition}{Target Latency}:
The target latency, denoted as $\tau$, is mathematically defined as:
\begin{gather}\scriptsize
\tau(\omega^t) = 
\left(\frac{Size(\mathcal{I}_j)}{\Lambda^u} + \rho^\mu(L) + \frac{Size(\mathcal{I}_j)}{\Lambda^u}\right) + \mathscr{C}(s_j) + \nonumber \\
\scriptsize
\left(\frac{eSize(\mathcal{O}_j)}{\Lambda^d} + \rho^\mu(L) + \frac{eSize(\mathcal{O}_j)}{\Lambda^d}\right)
\label{eq:target_lat}
\end{gather}
\noindent
where $\rho^\mu(L) = \underset{e_i \in \cal{E}}{median}(\rho_i^\mu(L))$ and $\mathscr{C}(s_j)$ is the median computation latency on an edge server when a certain number (say, $\nu$) of instances of the service is executed on it.  
\hfill$\blacksquare$
\end{definition}

It is important to note from Eq. \eqref{eq:target_lat} that to compute the transmission latency, we use only the user's uplink and downlink bandwidths. This is a reasonable assumption because, in general, the bandwidth of the edge server is significantly higher than that of the user device. As a result, when computing the minimum value between the user's uplink bandwidth and the edge server's downlink bandwidth, the user's uplink bandwidth typically provides the minimum value.

Since the actual size of the output parameters is not available during the computation of the target latency, we use the estimated size of the output parameters in the computation of the transfer time for the service output.

We choose the median propagation latency across all edge servers because each server, regardless of its propagation latency, has a non-zero probability of providing the least service latency. Therefore, instead of choosing the least propagation latency, we consider the median value across all servers. 

We select the value of $\nu$ based on profiling the computation time with varying numbers of parallel instances of the service. Initially, adding more parallel instances typically results in only minor increases in execution time. However, beyond a certain threshold, additional parallel instances cause a significant rise in execution time. We choose $\nu$ to strike a balance, ensuring sufficient parallelism to prevent a drastic increase in execution time while reflecting the fact that each edge server often handles multiple parallel instances.

We finally define the reward function.
\begin{definition}{Reward}:
The reward, denoted as ${\mathcal{R}}^{t+1}(\omega^t, {\mathscr{A}}^{t+1})$, is a function computed at timestep $t+1$ and is mathematically defined as:
\begin{equation}\scriptsize
{\mathcal{R}}^{t+1}(\omega^t, {\mathscr{A}}^{t+1}) = 
\begin{cases}
0 & \text{if } {\mathcal{L}}_R = \tau(\omega_t)\\
0 & \text{if } {\mathcal{L}}_R < \tau(\omega_t) \text{ and } |{\mathcal{E}}_k| = 1 \\
0 & \text{if } {\mathcal{L}}_R > \tau(\omega_t) \text{ and } |{\mathcal{E}}_k| = n \\
-\delta. e^{(n - |{\mathcal{E}}_k|)} & \text{if } {\mathcal{L}}_R > \tau(\omega_t) \text{ and } |{\mathcal{E}}_k| < n \\
-\delta. e^{({\mathcal{L}}_R - \tau(\omega_t))} & \text{if } {\mathcal{L}}_R < \tau(\omega_t) \text{ and } |{\mathcal{E}}_k| > 1 \\
\end{cases}
\label{eq:reward}
\end{equation}
where 
${\mathcal{E}}_k$ is the subset of edge servers chosen in action ${\mathscr{A}}^{t+1}$, and ${\mathcal{L}}_R$ is the latency achieved as a result of the action ${\mathscr{A}}^{t+1}$, calculated according to Eq. \eqref{eq:lat_redundancy}. The variable $n$ denotes the total number of edge servers and $\delta$ is an externally tunable positive real-valued hyperparameter used to adjust the value of ${\mathcal{R}}^{t+1}$, either scaling it up or down.

\hfill$\blacksquare$
\end{definition}

\begin{property}{Upper Bound}:
The upper bound of the reward function is $0$.
\hfill$\blacksquare$    
\end{property}

\begin{property}{Conditions to Achieve Maximum Reward}: The maximum reward is attained when:
\begin{itemize}[leftmargin=*]
    \item The achieved latency is exactly equal to the target latency irrespective of resource utilization.
    \item The achieved latency is less than the target latency, and the resource utilization is minimal.
    \item The achieved latency is higher than the target latency, but the resource utilization is maximum. It is important to note that, although our objective is not fulfilled in this scenario, SafeTail cannot select a better action to satisfy the objective under these conditions. \hfill$\blacksquare$  
\end{itemize}  
\end{property}

\begin{property}{Conditions to Obtain Negative Reward}: A negative reward is attained when:
\begin{itemize}[leftmargin=*]
    \item The achieved latency is higher than the target latency, however, there is a scope to increase the redundancy. In this case, the reward is determined according to the fourth case of Eq. \eqref{eq:reward}, where it is inversely proportional to the potential increase in redundancy that remains achievable.
    \item The achieved latency is lower than the target latency; however, there is still potential to reduce redundancy. In this case, the reward is chosen in the fifth case of Eq. \eqref{eq:reward}, where it is inversely proportional to the duration by which the target latency exceeds the achieved latency.\hfill$\blacksquare$
\end{itemize}    
\end{property}

\begin{property}{Characteristics of Reward Function}: In designing the reward function, we prioritize meeting the target latency over reducing resource utilization. This is reflected in the fourth and fifth cases of Eq. \eqref{eq:reward}. Notably, when the achieved latency exceeds the target and there is potential to increase redundancy, the penalty is more significant compared to situations where latency requirements are met but with higher resource usage. This characteristic of the reward function effectively optimizes not only the median latency but also the tail latency.
\hfill$\blacksquare$    
\end{property}

We now discuss the working principle of SafeTail. Fig. \ref{fig:SafeTail} (a) illustrates the overall structure of the framework. SafeTail employs a reward-based deep learning framework, with a feed-forward neural network with back propagation (FNN) as its core component, as presented in Fig. \ref{fig:SafeTail} (b). The FNN takes the state $\omega^t$ as input, therefore, the number of nodes in the input layer corresponds to $|\omega^t|$. The number of nodes in the output layer equals to the number of possible actions, which is $2^n - 1$.

The reward-based learning network in SafeTail is trained on an episode-wise basis. At the start of each episode, we randomly initialize the load, i.e., the number of active users on each edge server, and then begin the simulation. Each episode is divided into a certain number of steps (denoted as $\alpha_{s}$). In each step, a user requests a service execution, and the SafeTail framework selects a set of edge servers to execute the service, using either exploration or exploitation with a probability $\epsilon$ as described in Eq. \ref{eq:action}. Once the user receives the service response, a reward is calculated based on the actual service latency and the target latency heuristically computed for that service. The state-action-reward $(\omega^t, {\mathcal{A}}^{t+1}, {\mathcal{R}}^{t+1})$ tuple from each step is stored to train our reward-based learning framework, which is updated after every interval of $\kappa$ steps.

The reward ${\mathcal{R}}^{t+1}$ in the $(\omega^t, {\mathcal{A}}^{t+1}, {\mathcal{R}}^{t+1})$ tuple is translated to represent the target vector ${\mathcal{V}}^t$ for the FNN, which is then used to calculate the loss function for training the FNN. The target vector has a length equal to the number of output nodes in the FNN, i.e.,  $2^n -1$.
In typical classifiers, the target vector is a one-hot vector. However, in our framework, due to the lack of prior knowledge about the appropriate action for a given state, our target vector is not a one-hot vector. Instead, we ensure that the sum of all elements in the target vector equals 1. 
If the reward ${\mathcal{R}}^{t+1}$ for a particular action ${\mathcal{A}}^{t+1}$ is zero, the corresponding element of $A_k$ (assuming ${\mathcal{A}}^{t+1} = A_k$) in ${\mathcal{V}}^t$ is set to 1, and the rest of the elements are set to 0.
For other cases where ${\mathcal{R}}^{t+1}$ is negative, we start by initializing all elements of ${\mathcal{V}}^t$ with an equal value, i.e., $\frac{1}{2^n - 1}$. We then adjust the values for each element in ${\mathcal{V}}^t$.
The values corresponding to $A_k$ and all other actions $A_j$ that involve edge servers ${\mathcal{E}}_j$, where ${\mathcal{E}}_j$ is a subset of ${\mathcal{E}}_k$, which correspond to $A_k$, are  set as follows:
\begin{equation}   
     {\mathcal{V}}^t(j) = \max \left(0, \left(\frac{1}{2^n - 1} + {\mathcal{R}}^{t+1}\right)\right), ~~ \forall j \text{ where } {\mathcal{E}}_j \subseteq {\mathcal{E}}_k
\end{equation}
\noindent
Once all the elements in ${\mathcal{V}}^t$ corresponds to the actions $A_j$, $\forall j \text{ where } {\mathcal{E}}_j \subseteq {\mathcal{E}}_k$, are adjusted, the remaining value, i.e., $\left(1 - \sum \limits_{\forall j, {\mathcal{E}}_j \subseteq {\mathcal{E}}_k} {\mathcal{V}}^t(j)\right)$, is equally distributed among the remaining elements of ${\mathcal{V}}^t$.

Our reward-based learning network is trained for a predefined number of episodes (denoted as $\alpha_e$) until the training process is terminated.

In the next section, we discuss our experimental setup and analyze the performance of SafeTail.

%% file: 4Implementation.tex
\section{Implementation \& Experimental Analysis}

\subsection{Experimental Setup}
\noindent 
Our simulator was built using YAFS \cite{yafs} to model the topology of edge servers and user devices. We modified YAFS to be compatible with Python 3.11, incorporated uncertainties in the wireless network and execution time, enabled service scheduling on duplicate edge servers, and integrated our SafeTail framework to allow a user to schedule all the services they wish to invoke. We implemented our reward-based deep learning model using Keras with the TensorFlow backend (Keras v3.0.5 and TensorFlow 2.16.1).

\begin{figure}
\includegraphics[trim={10cm 4.5cm 0 0.7cm},clip,width=0.5\textwidth]{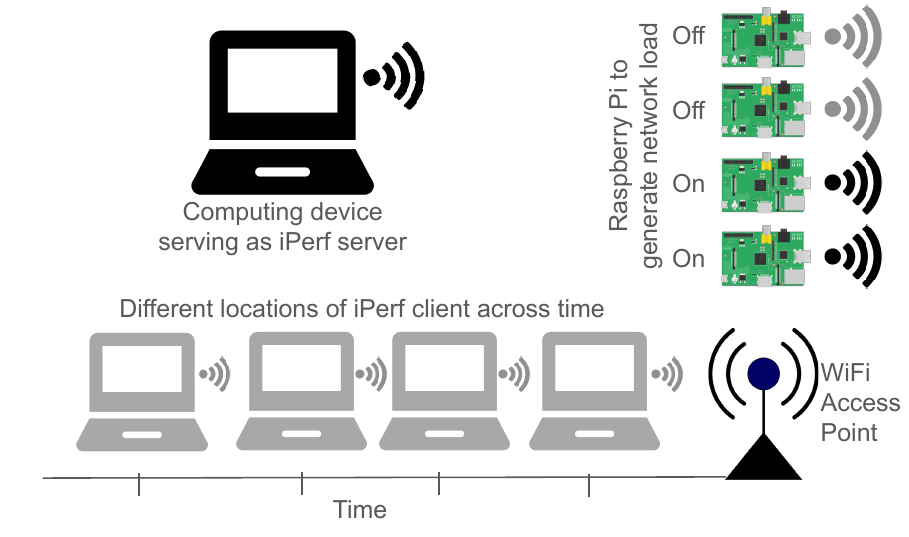}
    \vspace{-0.25in}
    \caption{Our experimental setup with two personal computing devices, four Raspberry Pi's, and one access point.}
    \label{fig:testbed-setup}
\end{figure}

To make our experiments realistic, we collected both network traces and execution traces, as illustrated below:

\noindent\textbf{Modeling Transmission Latency}: 
We first collected traces of network data over WiFi to obtain the distribution for the uplink and downlink bandwidths of both the server and client. To do this, we set up an access point (AP) and connected two personal computing devices to it via WiFi, as shown in Fig. \ref{fig:testbed-setup}. One device acted as the iPerf3 server, while the other served as the client. We collected network throughput data (in bits per second) reported by iPerf3, repeating the experiment 1000 times to gather each throughput value. To introduce additional network congestion, we added 1-4 Raspberry Pi (Model 3B) devices to the same WiFi network, simulating the presence of additional users. The varying number of Raspberry Pis captured the uncertainty in the uplink and downlink bandwidths of the edge servers. 
Since our framework assumes homogeneous edge server settings, a single experiment involving one edge server can be used to model the initial uplink and downlink bandwidths for all edge servers in the network. However, as the load on each edge server changes over time, the uplink and downlink bandwidths per user may vary at later timesteps.

\noindent\textbf{Modeling Propagation Latency}: 
In a similar networking setup, we sent 500 ping packets from one computing device to another to model propagation latency. We repeated this experiment by placing the client device at varying distances from the AP to capture propagation latency under different signal strengths, recording the round trip times reported by ping for each packet.

\noindent\textbf{Modeling Computation Latency}: 
To collect the execution traces, we run 1-5 instances of the chosen service in parallel on an Intel Core i7-11700 processor (8 cores at 2.5 GHz base frequency) and 16 GB RAM. For each instance, we note down the execution time of the service using the time command.

We selected three distinct services to evaluate SafeTail's performance: (i) single-shot object detection on images using YOLOv5 \cite{yolov5}, (ii) instance segmentation on images \cite{instance}, and (iii) noise removal from audio signals \cite{noise}. 
These services are integral to latency-sensitive autonomous systems, underscoring the importance of optimizing their tail latencies. The chosen services span computation times from hundreds of milliseconds to several seconds, showcasing SafeTail's utility across a range of scenarios. Additionally, these services cover two types of input modalities: images and audio.
We tested using benchmark datasets: the COCO dataset for images \cite{cocodataset}, and a Kaggle benchmark for audio files \cite{audio-dataset}.

For each service, we identified the characteristics of the input parameters that influence computation latency by analyzing its correlations with various parameters. For example, we examined the relationship between execution time and parameters like file size, resolution/duration, and the number of processes. In image-based services, resolution is the relevant parameter, whereas duration applies to audio noise removal.

\begin{figure}[!t]
 \centering
 \includegraphics[width=0.3\linewidth]{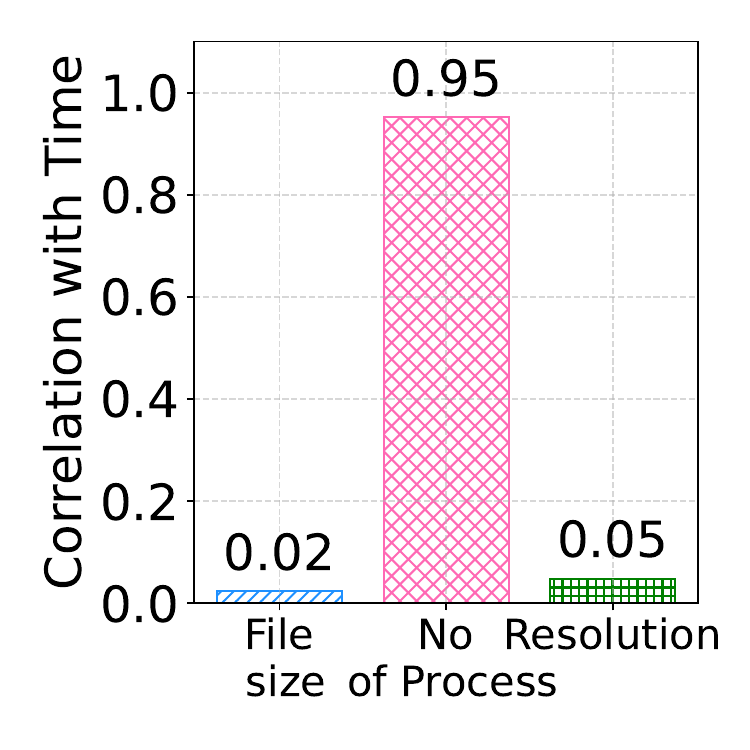}
\includegraphics[width=0.3\linewidth]{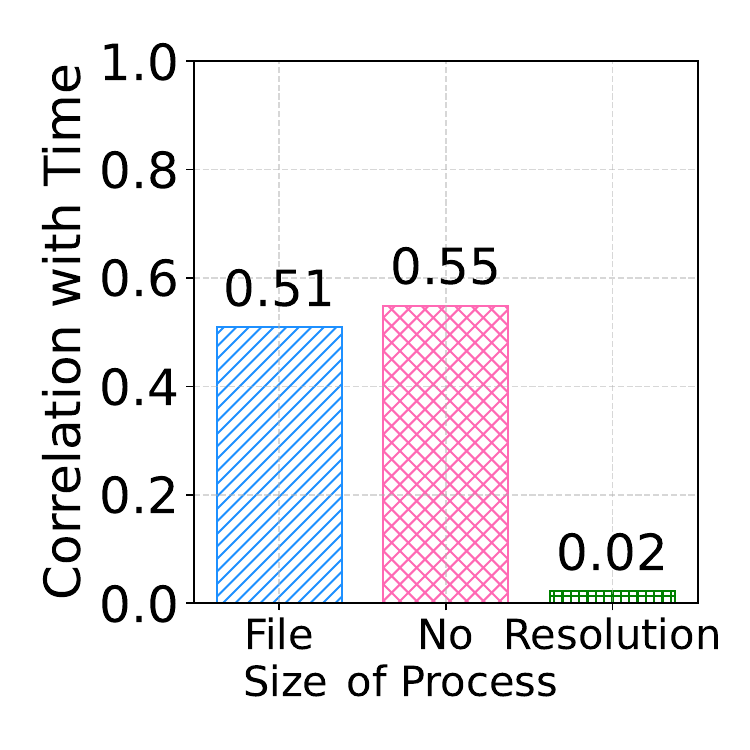}
\includegraphics[width=0.3\linewidth]{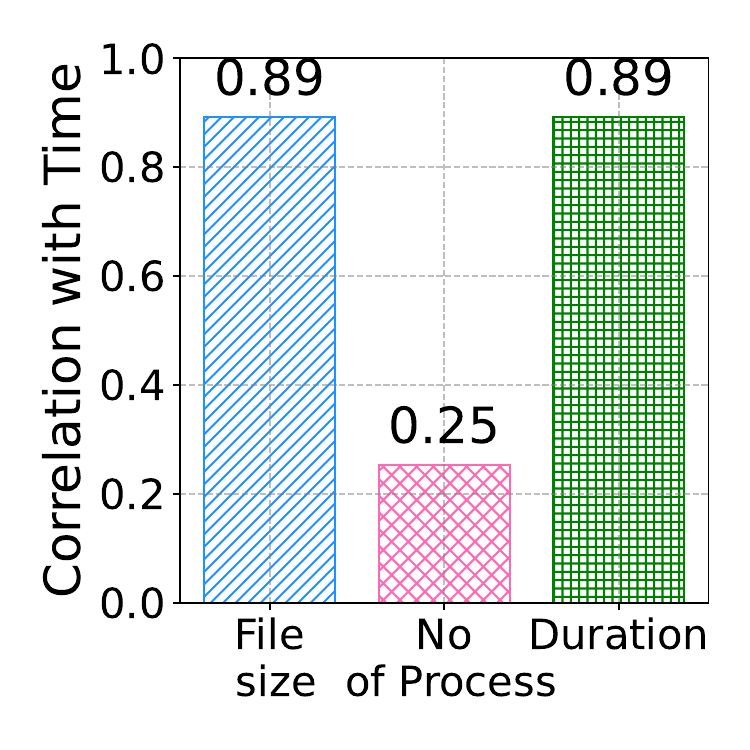}\\
(a) ~~~~~~~~~~~~~~~~~~~(b)~~~~~~~~~~~~~~~~~~~(c)
     \caption{Correlation between computation latency and different input parameters for (a) YOLOv5, (b) Instance segmentation, (c) Audio noise removal}
     \label{fig:corr}
 \end{figure}

We first prepared a dataset with file sizes, computation latency, and resolutions, converting the latter into numerical formats for consistency. We then calculated the Spearman correlation coefficients to quantify these relationships and visualized them using bar plots, as shown in Fig. \ref{fig:corr}.

In the case of object detection, we found that execution time is strongly correlated only with the number of processes, while file size and resolution showed no significant correlation. For instance segmentation, both file size and the number of processes had moderate correlations with execution time, but resolution exhibited a minimal correlation. Similarly, for noise removal from audio signals, there was a strong correlation with file size and duration and a moderate correlation with the number of processes. Parameters with small correlations were discarded, while those with stronger correlations were retained for further analysis.

After identifying all the parameters that influence the computation latency of a service, we created a dataset by executing each service on the benchmark test dataset. During each execution, we recorded the values of the selected parameters. This dataset was then used to train our regression model.

These datasets are subsequently used to formulate a regression model capable of predicting the execution time for each service. We employed a multilayer perceptron for this purpose, where computation latency serves as the response variable, and the selected parameters for each service act as the regressors. 

To introduce uncertainty, we implemented the following strategy. For each number of processes (i.e., 1 to 5 processes that are executed in parallel) of a service, we obtained the standard deviation, denoted as $\sigma_k$ (where $k$ is the number of parallel processes), of the computation latency. After the regression model generates an output, denoted as $\hat{y}$, for $k$ parallel processes, we further introduced Gaussian noise with a mean of $\hat{y}$ and a standard deviation of $\sigma_k$.

\noindent\textbf{Modeling Load on Each Edge Server}: We utilized the dataset \cite{li2021profit} to model the load variations for each base station. Starting with an initial number of active users on each edge server, we then adjusted the load according to the dataset's distribution.

%% file: 5Results.tex
\balance
\subsection{Baselines}
\noindent
We implemented and evaluated four baseline methods, each employing different strategies to optimize latency. We compared the total elapsed time to complete the service by each baseline method with SafeTail. For a fair comparison, these methods also incorporate redundancy. The four strategies are:

\emph{(i) Oracle (Optimal)}: This method allows a service to execute on all edge servers connected to the user's device. The recorded latency is the minimum observed, i.e., the latency of the fastest responding edge server. While this approach guarantees the minimal possible latency by maximizing redundancy, it is impractical as it increases network congestion and resource utilization.

\emph{(ii)  Random (Rand-x)}: This method involves using one or more randomly selected edge servers for service execution, depending on the allowed redundancy. In our experiment, we implement three sub-methods: Rand-1, Rand-2, and Rand-3, where the number indicates the level of redundancy.

\emph{(iii)  Edge Server with Minimum Propagation Latency (MinProp-x)}: This method directs the service to one or more edge servers with the lowest propagation latency, depending on the allowed redundancy. Here also, x represents the level of redundancy.
    
\emph{(iv)  Edge Server with Minimum Load (MinLoad-x)}: This method sends the service to one or more edge servers that are running the fewest number of services, with the number of edge servers selected based on the redundancy level x.

\subsection{Performance Metrics}
\noindent
The following metrics are used to measure the performance of SafeTail in comparison to the baseline methods.

\emph{(i)} \textbf{Access Rate:} The access rate of a method is defined as the ratio between the number of redundant edge servers selected for service execution and the total number of edge servers reachable from the user device.
    
\emph{(ii)}  \textbf{Latency Deviation:} The latency deviation is the difference between the target latency and the achieved latency.

In our experimental analysis, we present the results using the following visualizations for each use case:

\emph{(i)}  Average access rate versus average episode number: This visualization illustrates how the access rate fluctuates in response to the dynamic state of edge servers over time.  

\emph{(ii)}  Average latency deviation versus average episode number: This visualization demonstrates the characteristics of latency deviation as the number of episodes increases.

\emph{(iii)}  Average absolute value of reward versus average episode number: This visualization illustrates that the absolute value of the reward decreases and eventually stabilizes as the number of episodes increases. 

\emph{(iv)}  Comparison with baseline methods in terms of median and tail latency: This analysis compares SafeTail with baseline methods in terms of median and higher percentiles (\pni, \pnf, and \pnn) of latency, also known as tail latency. We first compare the latency achieved by SafeTail to the optimal latency. Next, we assess the average target latency relative to both the optimal latency and the latency achieved by SafeTail. Finally, we demonstrate that SafeTail outperformed all baseline methods in most cases.

In our analysis, we average each performance metric over $\beta$ steps and present the results. In our plots, the average episode number refers to the ratio of the total number of steps to the parameter $\beta$. 
In the plot of the average absolute value of reward versus episode number, we display the training reward (for 50 × $\beta$ steps) followed by the testing reward (for 20 × $\beta$ steps). Furthermore, in our comparison plots, we have included the average value of the target latency. 

\begin{table}[!t]
\centering
\caption{Model Configuration}
\begin{tabular}{l|c|c|c}
\hline
\textbf{Parameter} & \multicolumn{3}{c}{\textbf{Value}} \\ \hline
\multicolumn{4}{c}{Configuration of SafeTail}\\
\hline
$n$ & \multicolumn{3}{c}{5}\\
Maximum load for & \multicolumn{1}{c}{} & \multicolumn{1}{c}{\multirow{2}{*}{4}} & \multicolumn{1}{c}{}\\
each edge server & \multicolumn{1}{c}{} & \multicolumn{1}{c}{} &  \multicolumn{1}{c}{}\\
\hline

\multicolumn{4}{c}{Configuration of Reward-based Learning Framework}\\
\hline
& UC-1 & UC-2 & UC-3 \\ \hline

Learning rate & 1e-07 & 1e-06 & 1e-07 \\ 
$\delta$ & 0.003 & 0.0015 & 0.003 \\ 
$\epsilon_{\text{decay}}$ & $3.3 \times 10^{-6}$ & $3.3 \times 10^{-6}$ & $3.3 \times 10^{-6}$ \\ 
$\epsilon_{\text{min}}$ & $0.01$ & $0.01$ & $0.01$\\ 
$\alpha_s$ & 256 & 256 & 256\\ 
$\alpha_e$ & 50 & 30 & 50 \\ 
$\kappa$ & 128 & 128 & 128\\ 
$\beta$ & 200 & 120 & 200 \\ \hline

\multicolumn{4}{c}{Configuration of FNN} \\\hline
Number of hidden layers & \multicolumn{3}{l}{5} \\ 
Activation function & \multicolumn{3}{l}{ReLU} \\ 
Output function & \multicolumn{3}{l}{Softmax} \\
Loss function & \multicolumn{3}{l}{Categorical Cross entropy}\\ 
Optimization function & \multicolumn{3}{l}{Adam} \\ \hline 
\multicolumn{4}{r}{UC: Use case}
\end{tabular}
\label{tab:model_parameters}
\end{table}

\begin{figure*}[!t]
 \centering
 \includegraphics[width=0.24\textwidth]{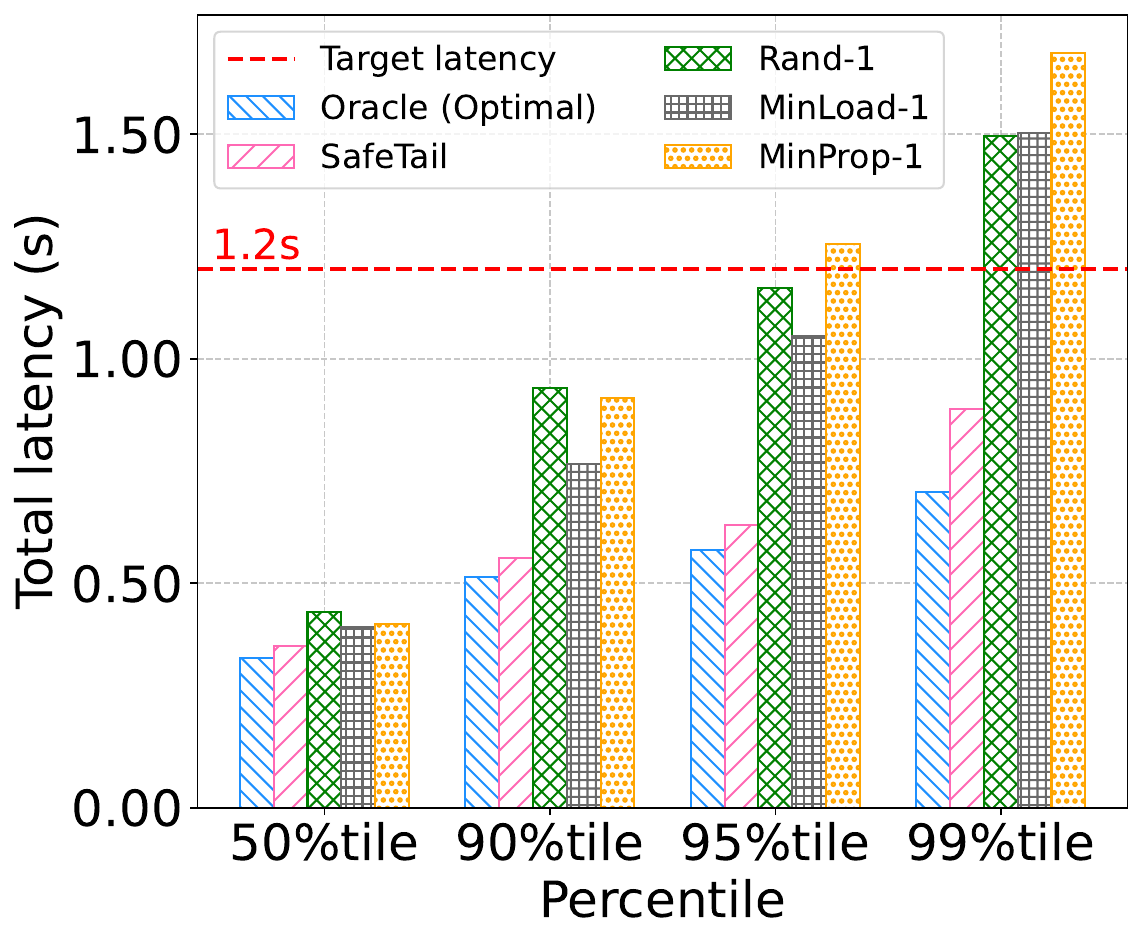}
\includegraphics[width=0.24\textwidth]{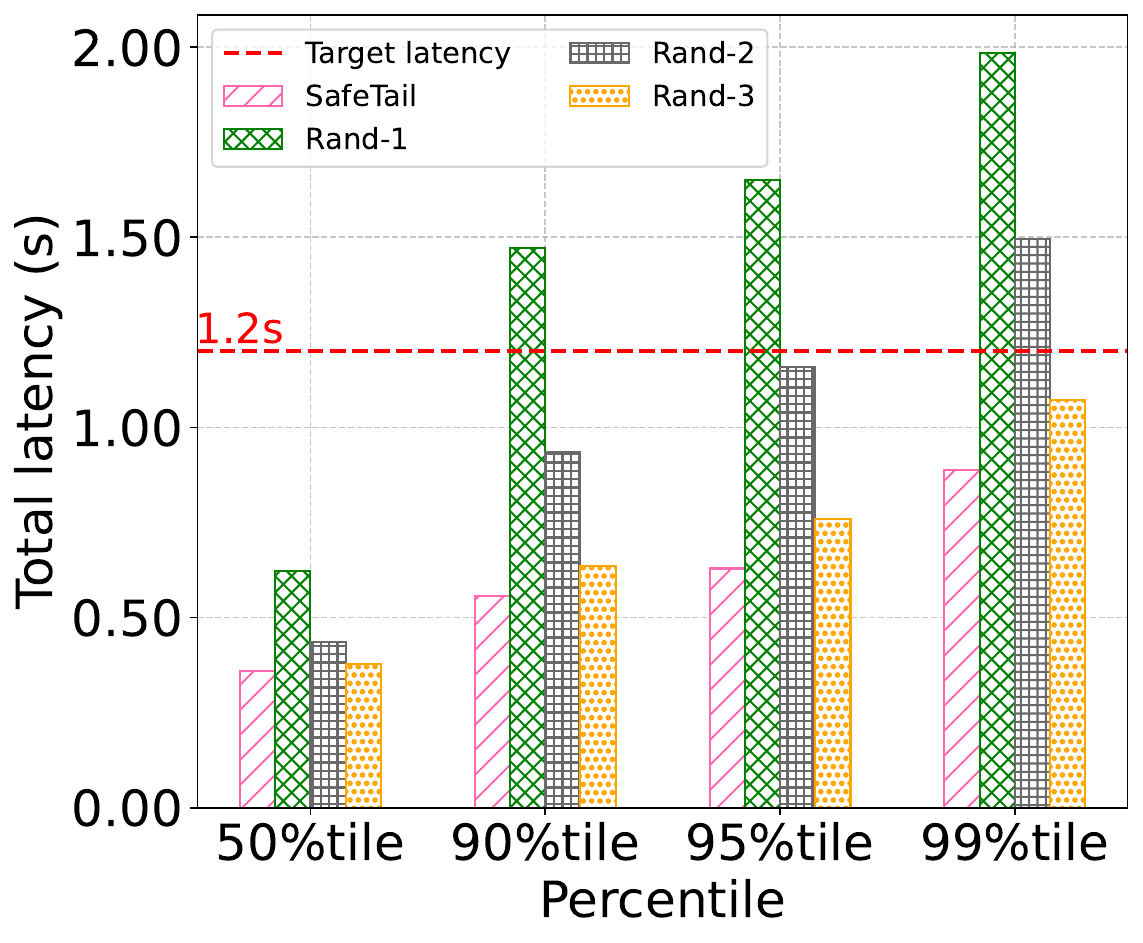}
\includegraphics[width=0.24\textwidth]{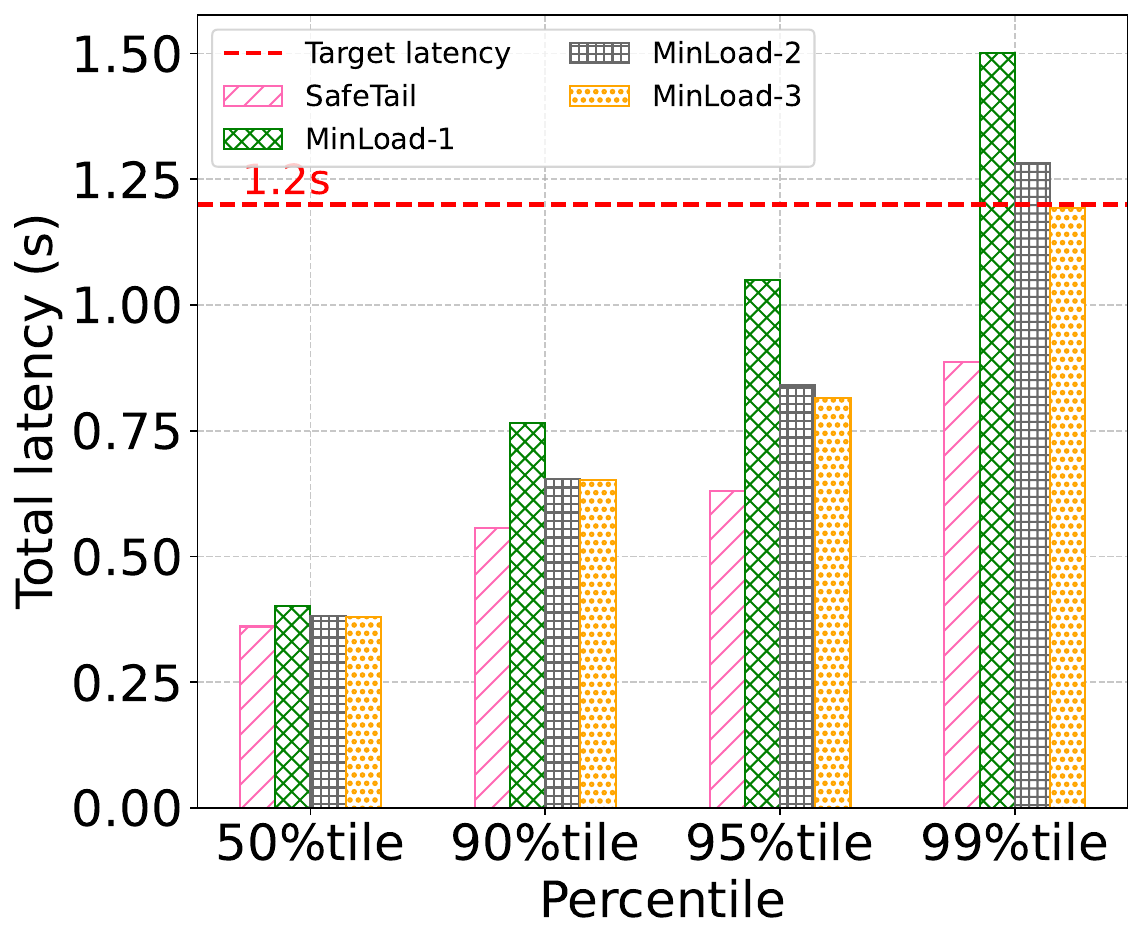}
\includegraphics[width=0.24\textwidth]{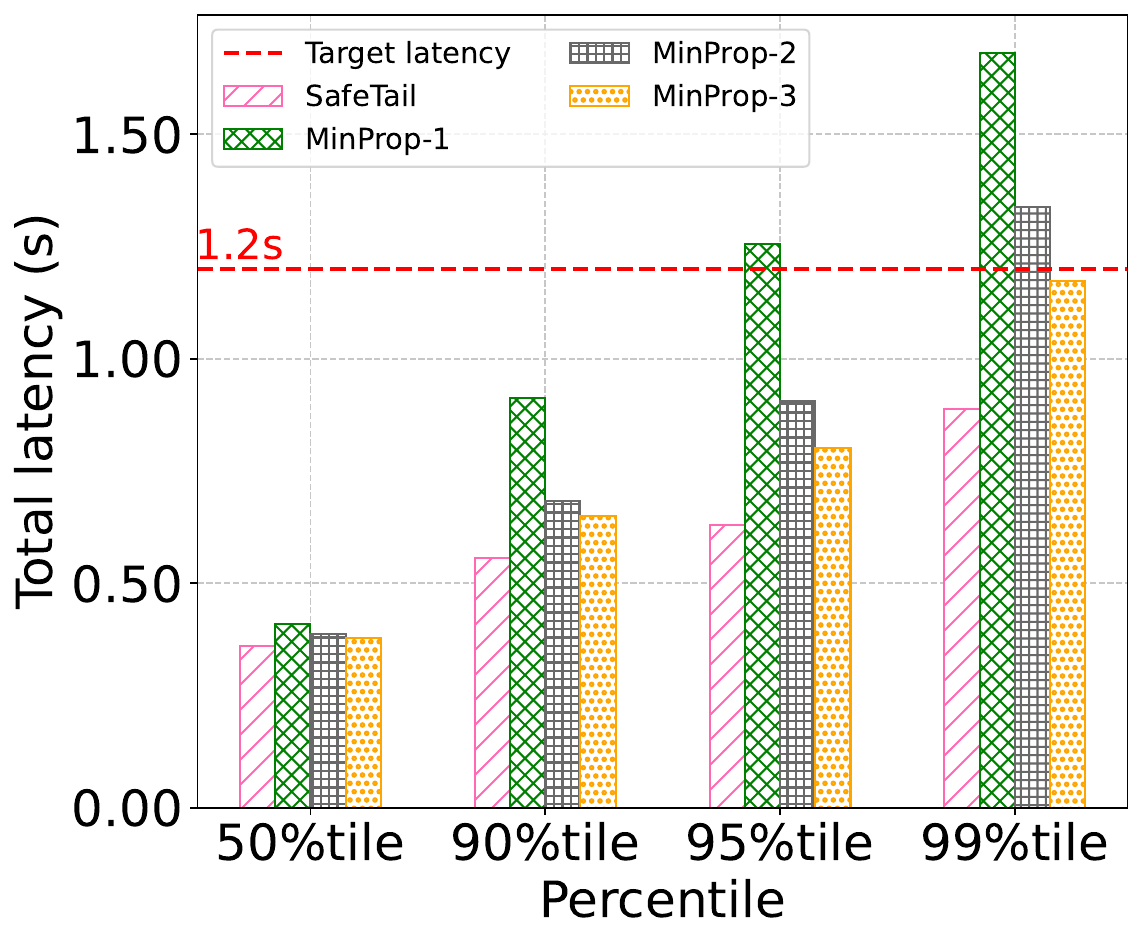}\\
(a)~~~~~~~~~~~~~~~~~~~~~~~~~~~~~~~(b)~~~~~~~~~~~~~~~~~~~~~~~~~~~~~~~~~~~(c)~~~~~~~~~~~~~~~~~~~~~~~~~~~~~~~~~(d)\\~\\
 (e) \includegraphics[width=0.24\textwidth]{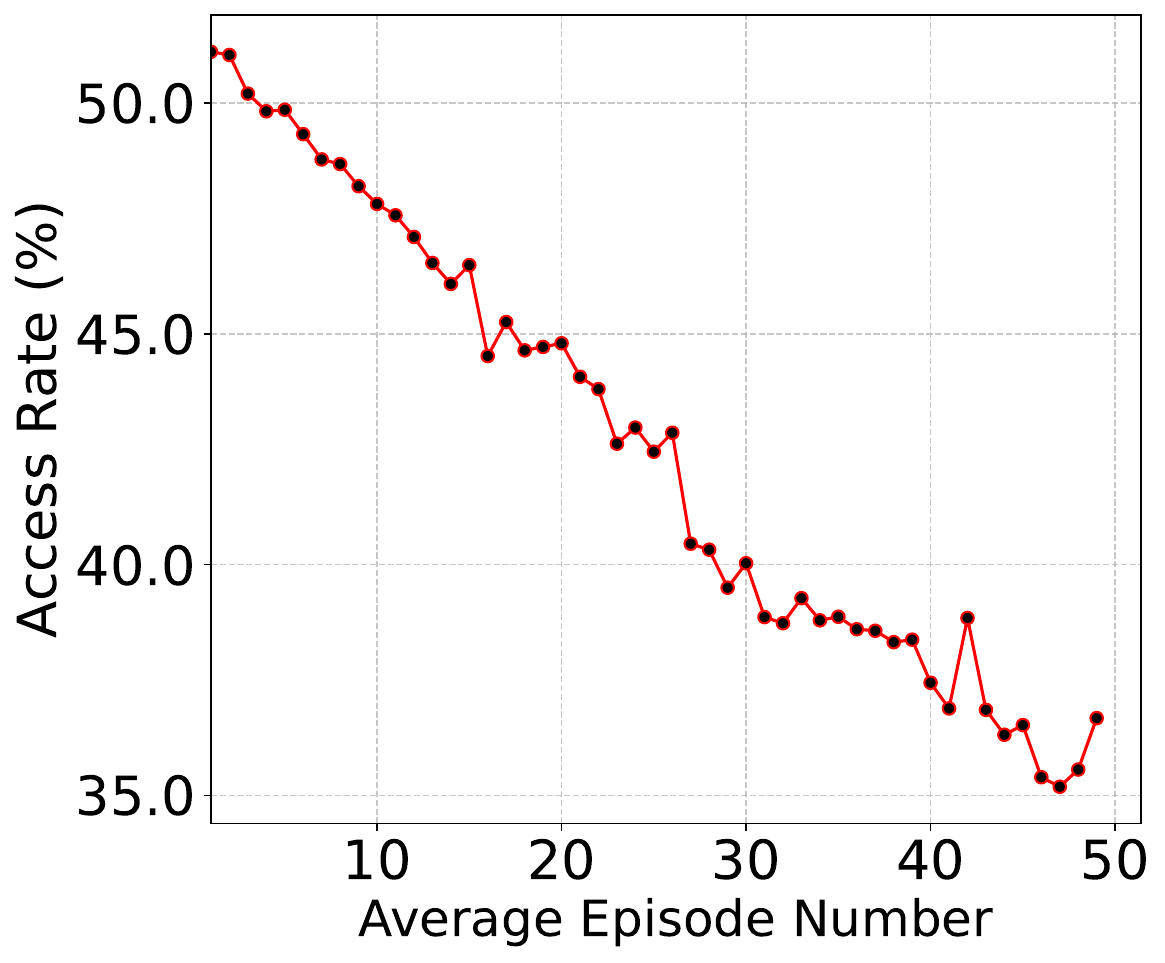}\quad
  (f) \includegraphics[width=0.24\textwidth]{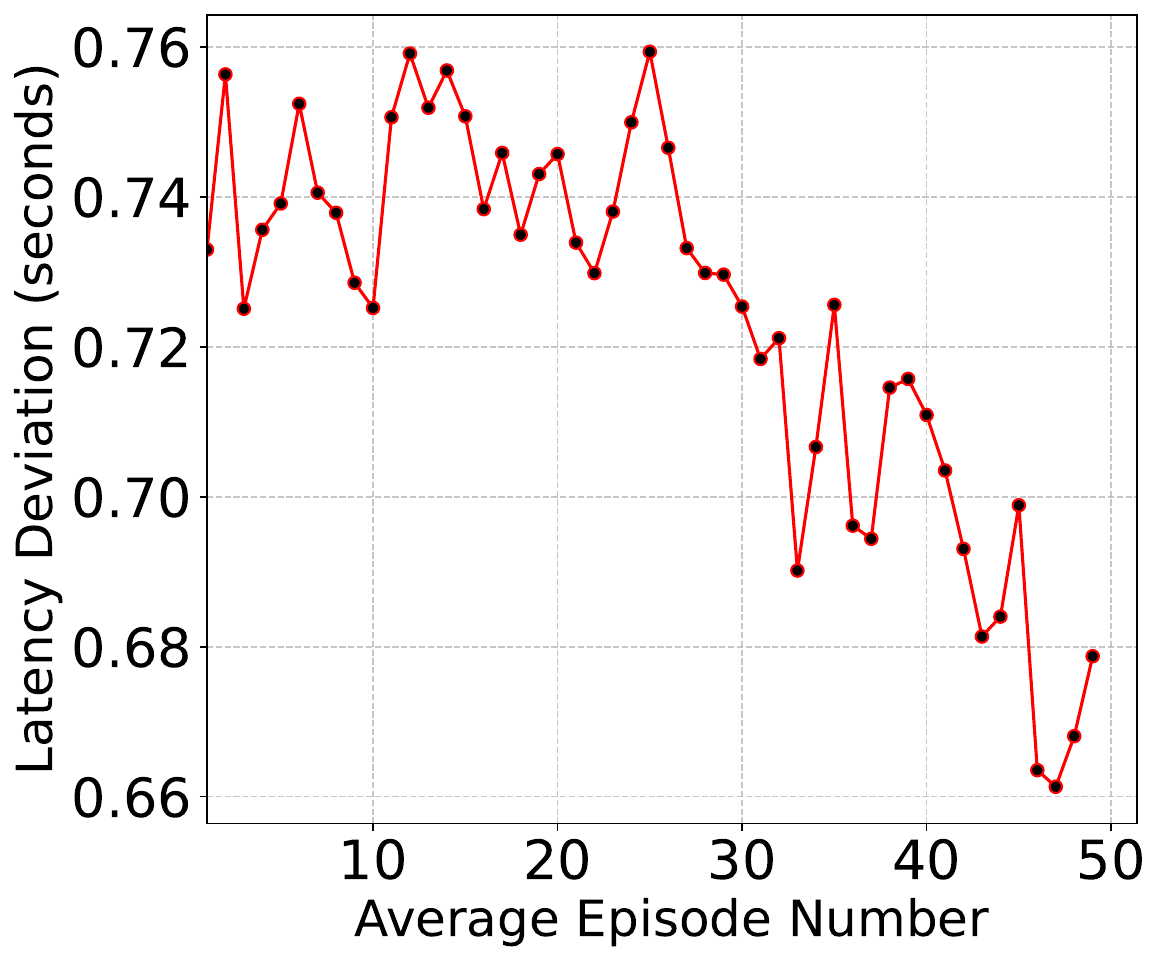}\quad
 (g) \includegraphics[width=0.24\textwidth]{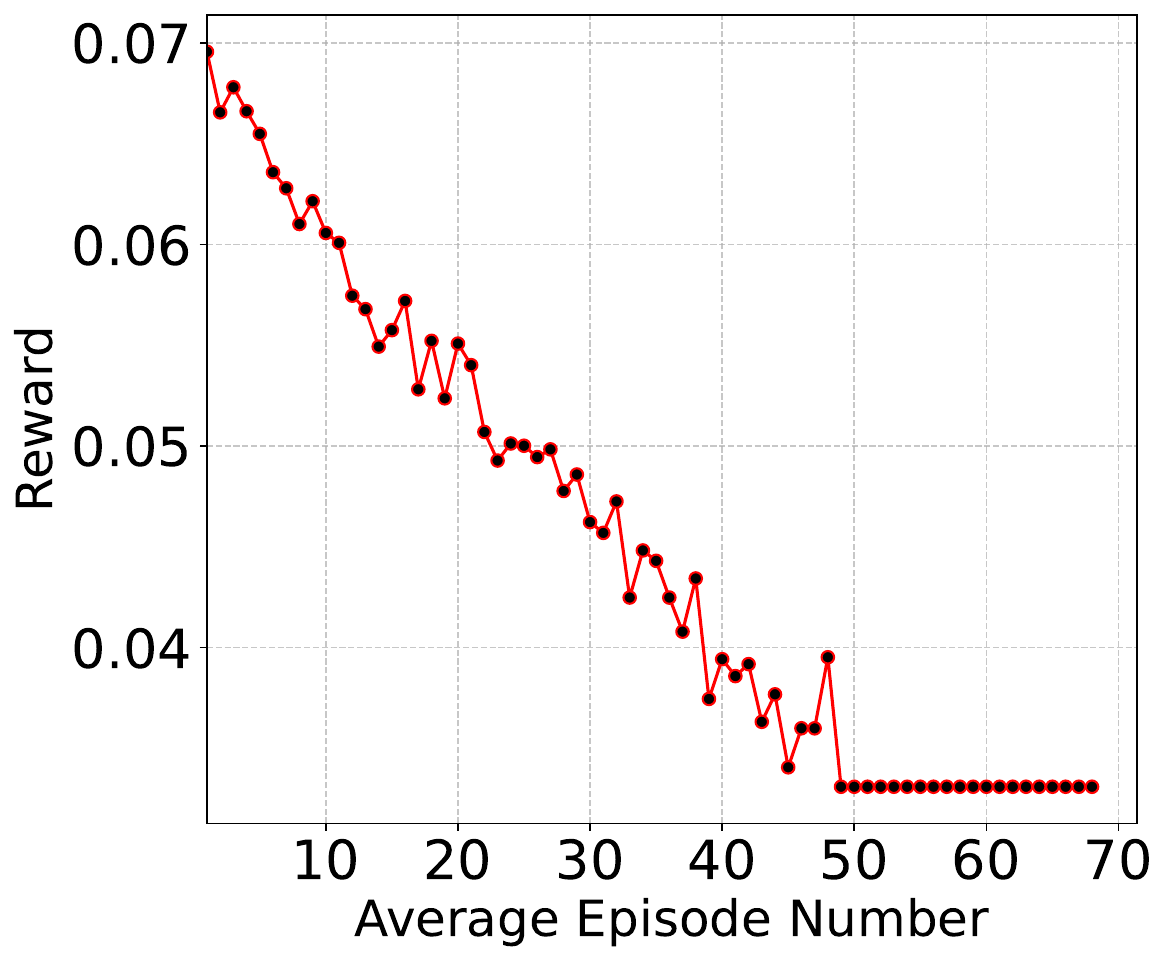}\\
  \caption{Use Case-1: (a) Latency comparison among SafeTail, target latency and baseline methods, (b) SafeTail versus Rand-x, (c) SafeTail versus MinLoad-x, (d) SafeTail versus MinProp-x, (e) Access rate of SafeTail, (f) Latency Deviation of SafeTail, and (g) Reward distribution of SafeTail.}
     \label{fig:yolo_comparison}
 \end{figure*}

\subsection{Configuration of the Parameters}
\noindent
Table \ref{tab:model_parameters} presents the configuration of the hyperparameters used in our framework. The values of the hyperparameters were determined experimentally using the validation dataset.

\subsection{Experimental Analysis}
\noindent
As discussed earlier, we now analyze the performance of SafeTail across the following three different use cases.

\subsubsection{Use Case-1: Object Detection using YOLOv5}
Fig. \ref{fig:yolo_comparison} illustrates the performance of SafeTail on the YOLOv5 service. Our observations are as follows:

\begin{enumerate}[leftmargin=*]
\item[(i)] Improvement in median latency: As shown in Fig. \ref{fig:yolo_comparison}(a), SafeTail attained a higher latency compared to Oracle's Optimal value, though with a small margin. 
Specifically, Oracle's Optimal value attained a speed-up in median latency of 1.08x compared to SafeTail. 
However, as evident from the figure, this median latency is 1.72x, 1.11x, and 1.13x faster than that of Rand-1, MinLoad-1, and MinProp-1, respectively.

\item[(ii)] Improvement in tail latency: As observed in Fig. \ref{fig:yolo_comparison}(a), even though SafeTail attained higher latency compared to Oracle's Optimal value by a small margin, it achieved significantly better tail latency compared to the baseline methods. For the \pni, \pnf, and \pnn percentiles, SafeTail experienced speed-ups in tail latencies of 0.93x, 0.92x, and 0.79x compared to Oracle's Optimal value, respectively. However, these tail latencies represent a speed-up of 2.64x, 2.62x, and 2.24x compared to Rand-1, 1.37x, 1.66x, and 1.69x compared to MinLoad-1, and 1.63x, 1.99x, and 1.89x compared to MinProp-1.

\item[(iii)] Achieved latency with respect to target latency: As evident from Fig. \ref{fig:yolo_comparison}(a), SafeTail consistently achieves latency below the target latency, including for tail latency cases.

\item[(iv)] Comparison with baselines with higher access rates: 
We further analyzed the performance of each baseline method by increasing redundancy up to 3, as shown in Fig. \ref{fig:yolo_comparison}(b)-(d). The results indicate that even with higher access rates, the latency values of all baseline strategies remain higher compared to SafeTail. This trend is consistent across all percentile values, demonstrating the effectiveness of SafeTail. Redundancy levels 4 and 5 were not considered in this experiment because level 5 is equivalent to Oracle's Optimal latency, and level 4 closely approximates it. Additionally, it may be noted the access rate of SafeTail varies between 1 to 3 edge servers.

\item[(v)] Variation in access rate: As noted from Fig. \ref{fig:yolo_comparison}(f), the latency deviation achieved by SafeTail gradually decreases and stabilizes with an increasing number of episodes. The reason for this trend is as follows: As observed in Fig. \ref{fig:yolo_comparison}(a), SafeTail consistently maintains latency below the target, even in tail latency scenarios. To achieve a lower access rate, as shown in Fig. \ref{fig:yolo_comparison}(e), SafeTail reduces the latency deviation while still keeping the latency value within the target limits, even if it means compromising slightly on the latency value.

\item[(vi)] Latency deviation: As noted from Fig. \ref{fig:yolo_comparison}(f), the latency deviation obtained by SafeTail gradually decreased with an increasing number of episodes.

\item[(vii)] Stabilization of reward function: We show in Fig. \ref{fig:yolo_comparison}(g) that the absolute value of the average reward curve eventually converges to a consistent value, indicating that SafeTail closely approximates the optimal latency and provides stable performance relative to the target latency.

\end{enumerate}

 \begin{figure*}[t]
 \centering
 \includegraphics[width=0.24\textwidth]{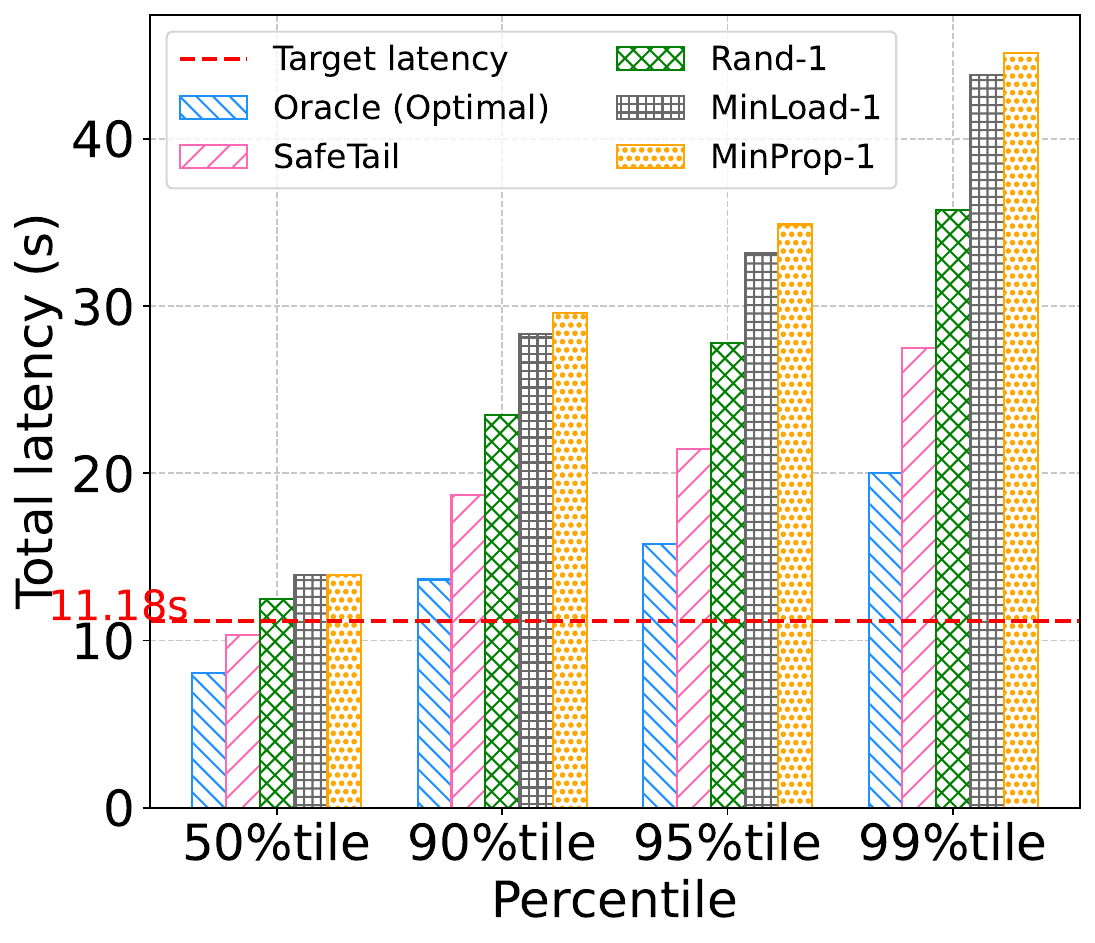}
\includegraphics[width=0.24\textwidth]{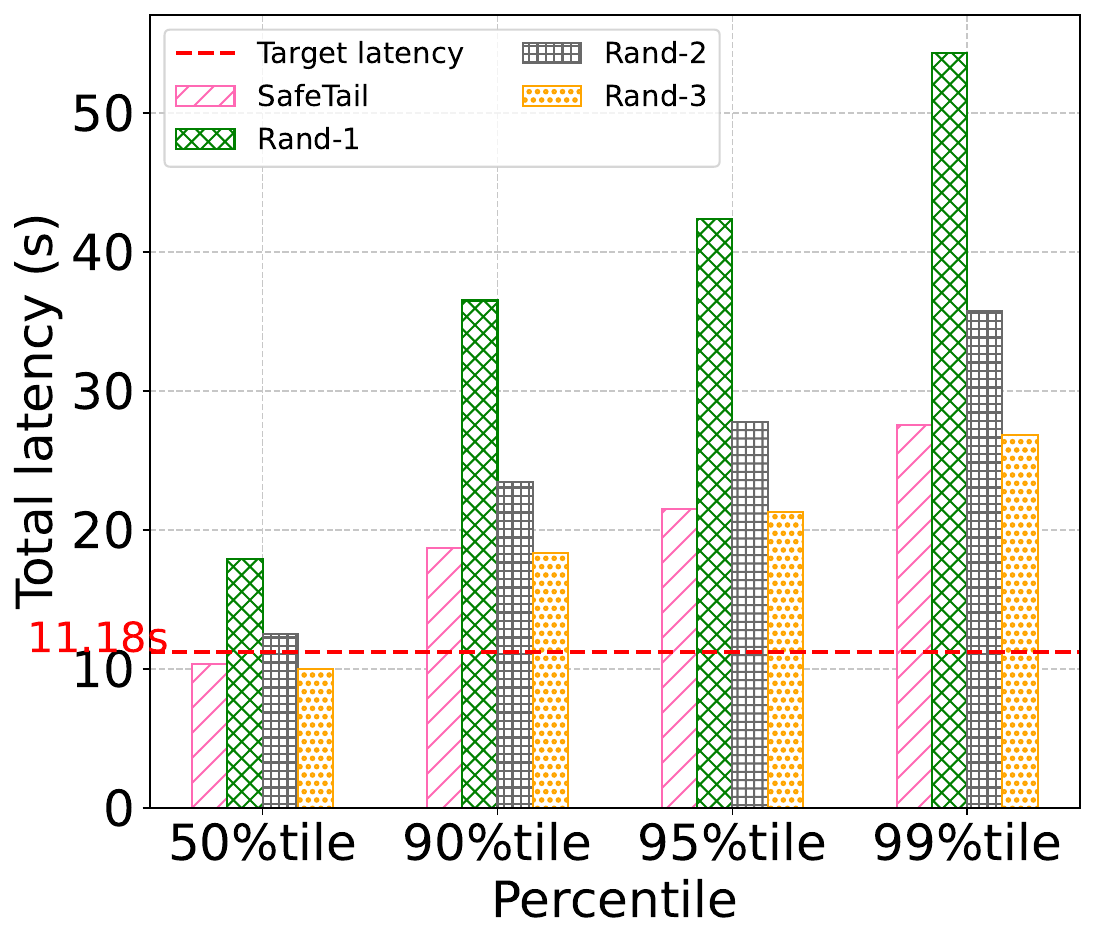}
\includegraphics[width=0.24\textwidth]{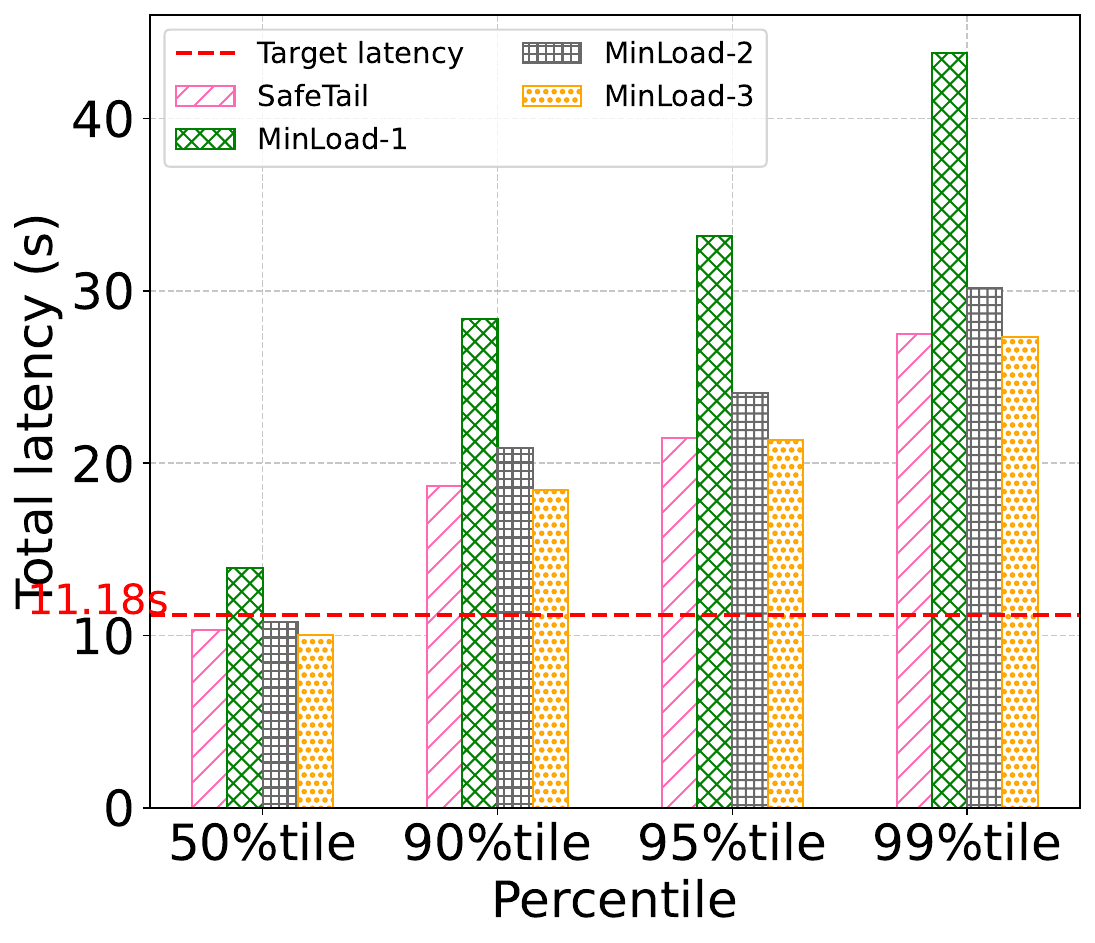}
\includegraphics[width=0.24\textwidth]{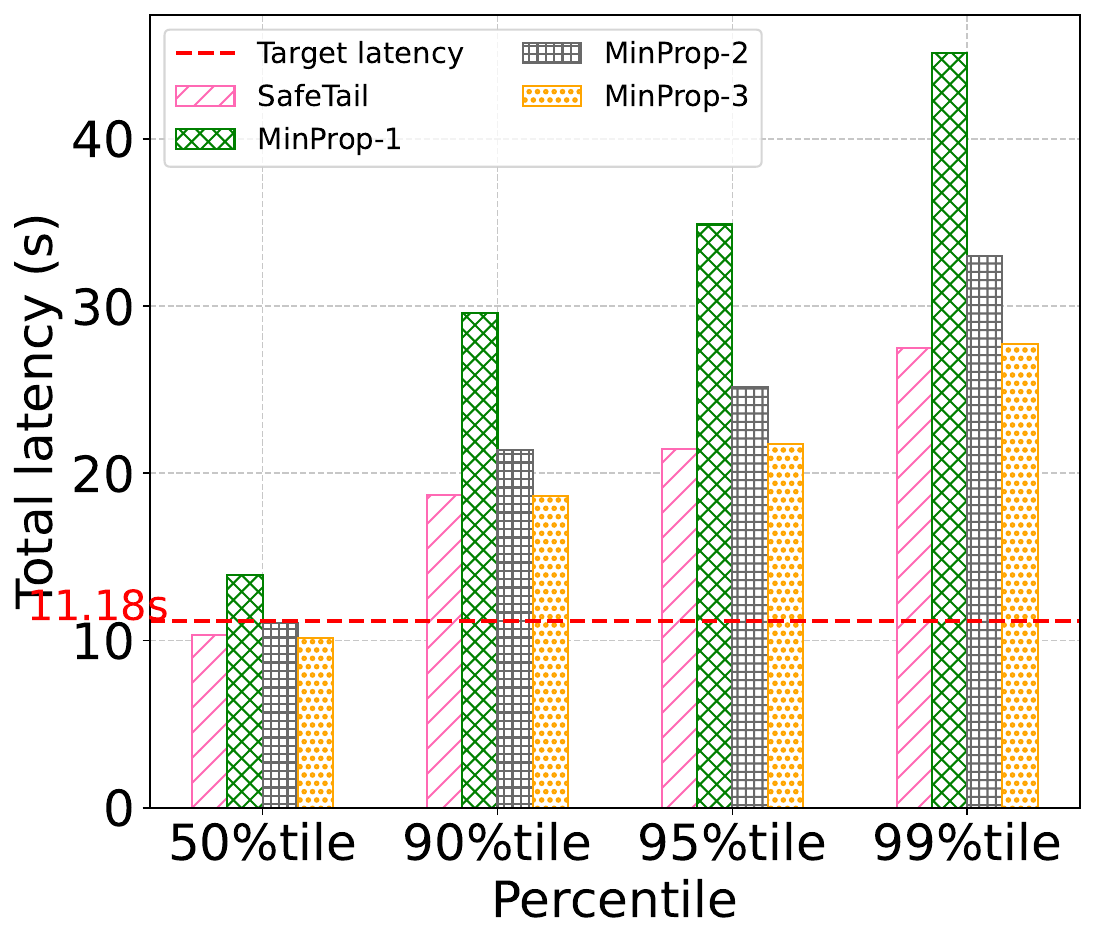}\\
(a)~~~~~~~~~~~~~~~~~~~~~~~~~~~~~~~(b)~~~~~~~~~~~~~~~~~~~~~~~~~~~~~~~~~~~(c)~~~~~~~~~~~~~~~~~~~~~~~~~~~~~~~~~(d)\\~\\
(e) \includegraphics[width=0.24\textwidth]{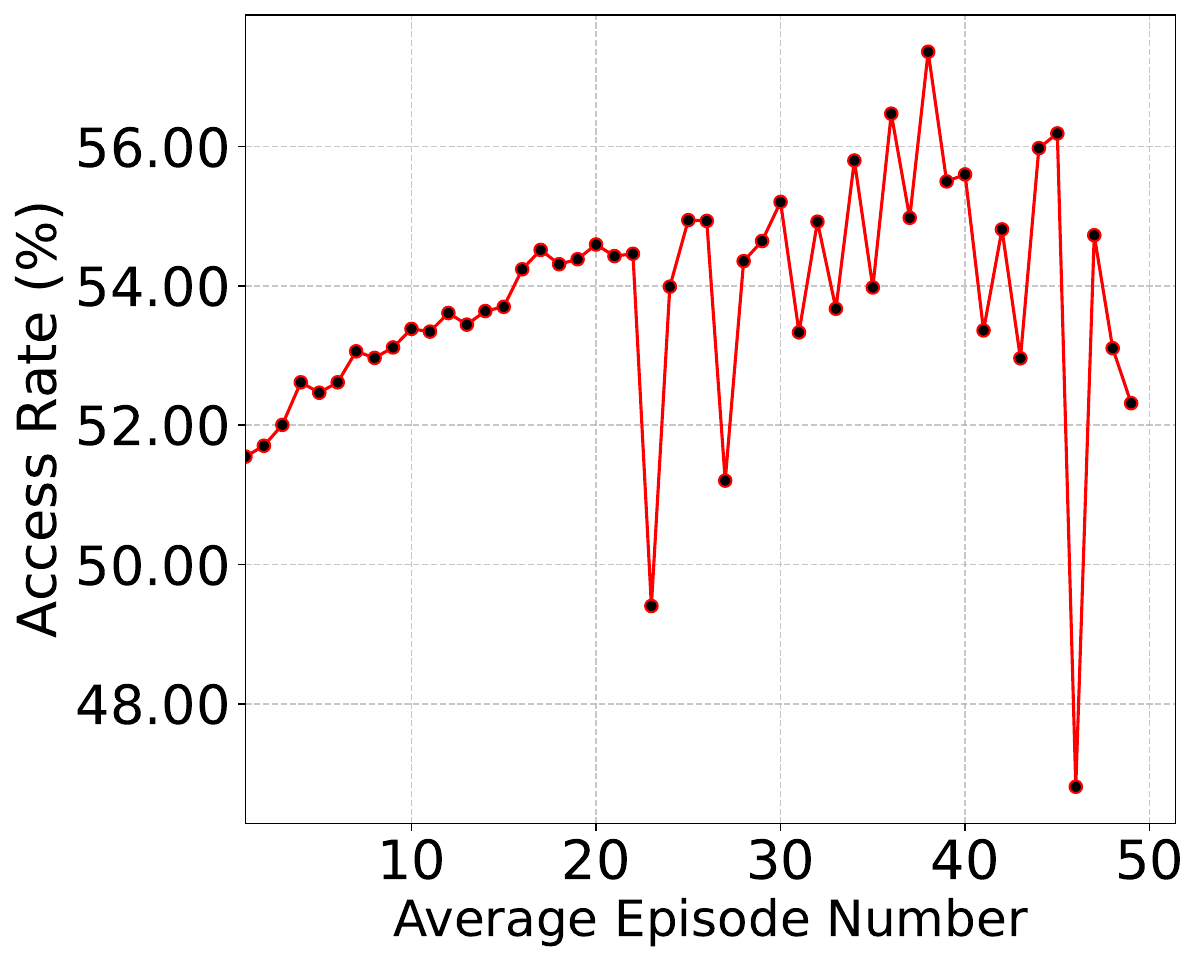}\quad
 (f) \includegraphics[width=0.24\textwidth]{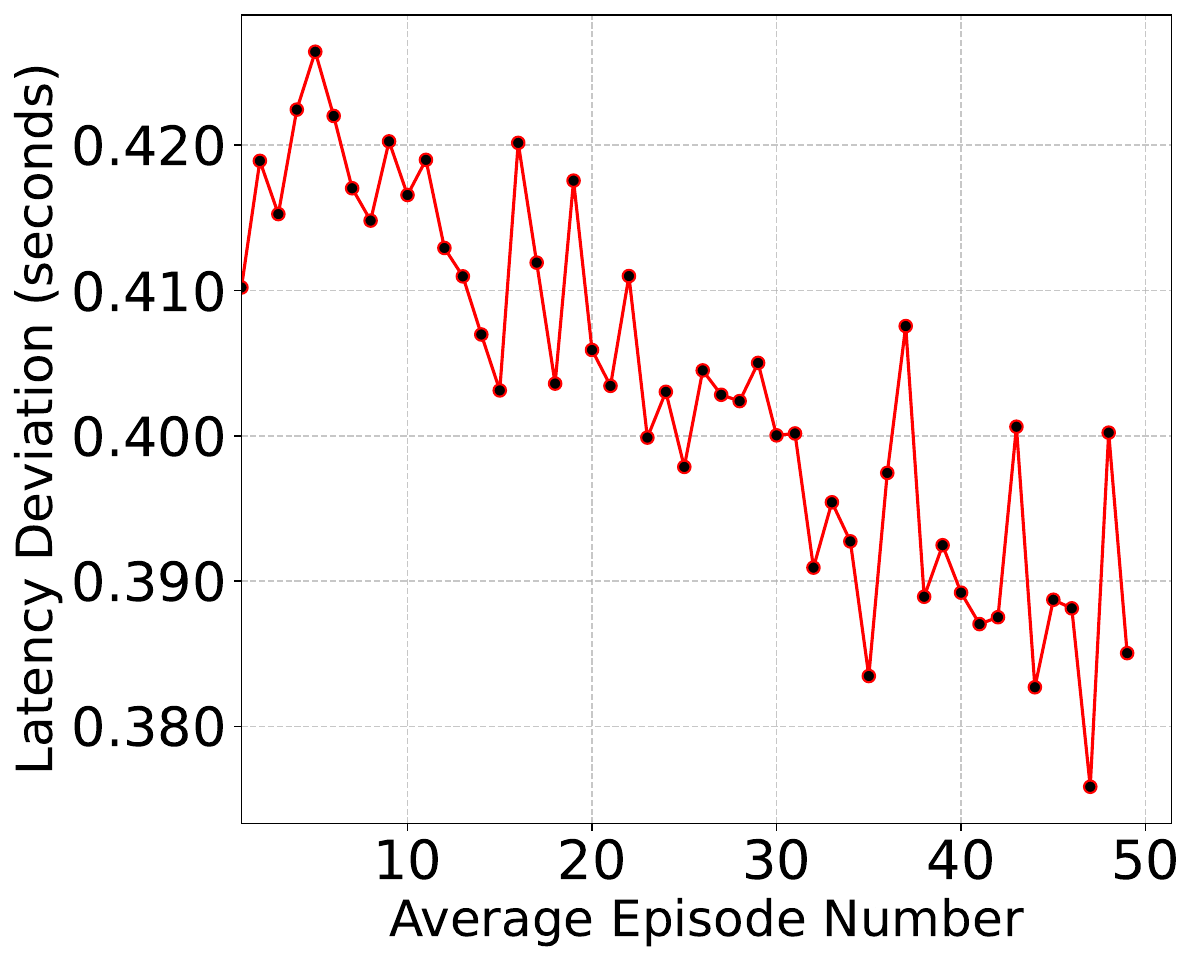}\quad
 (g) \includegraphics[width=0.24\textwidth]{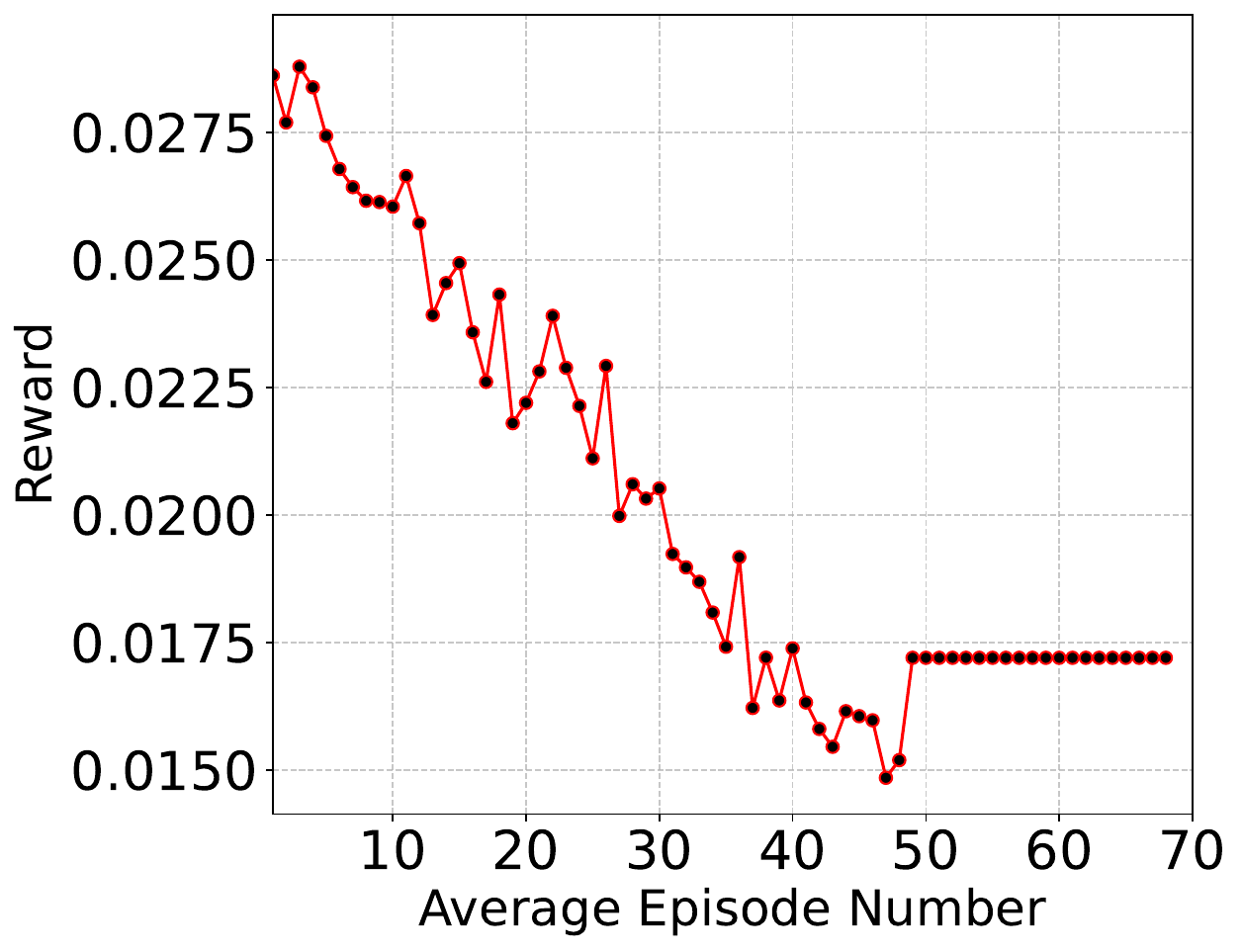}
  \caption{Use Case-2: (a) Latency comparison among SafeTail, target latency and baseline methods, (b) SafeTail versus Rand-x, (c) SafeTail versus MinLoad-x, (d) SafeTail versus MinProp-x, (e) Access rate of SafeTail, (f) Latency Deviation of SafeTail, and (g) Reward distribution of SafeTail.}
     \label{fig:ins_seg_comparison}
 \end{figure*}

  \begin{figure*}
 \centering
 \includegraphics[width=0.24\textwidth]{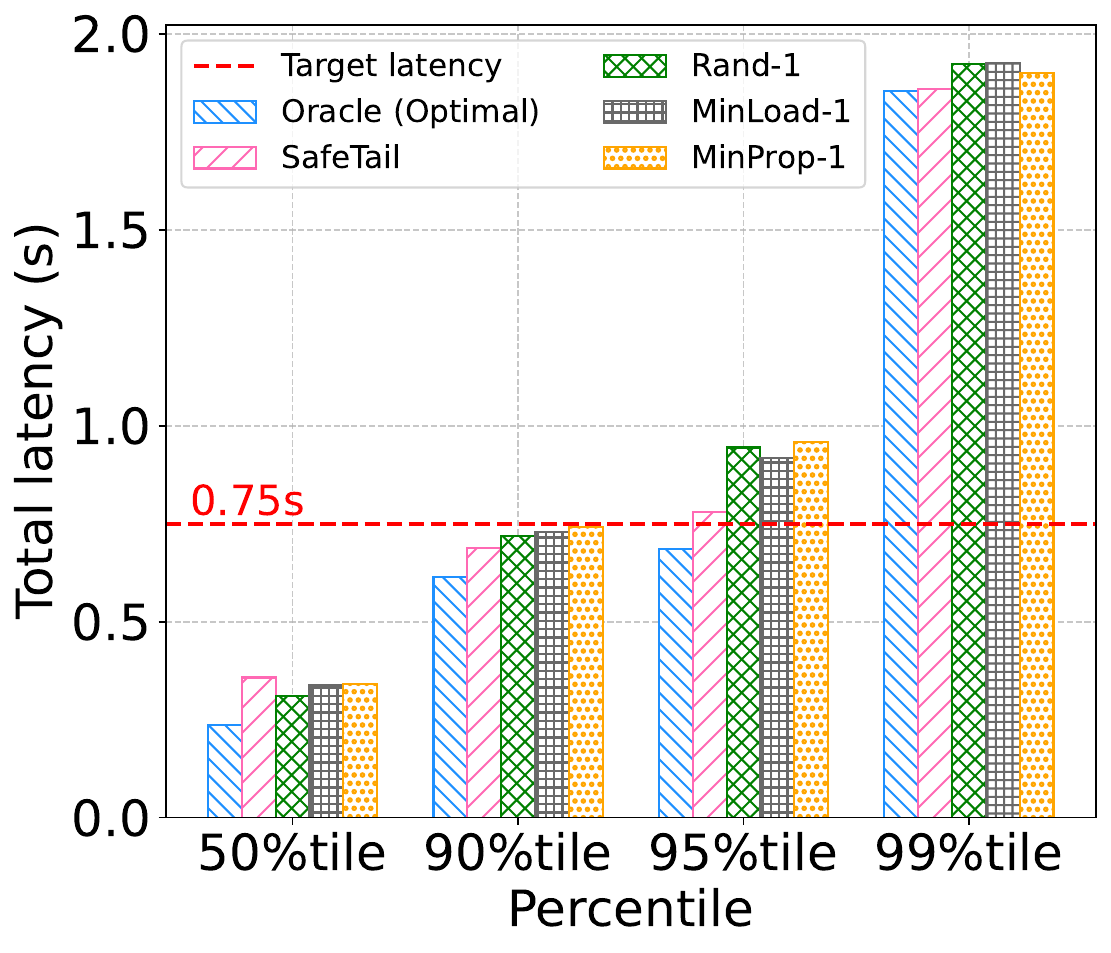}
\includegraphics[width=0.24\textwidth]{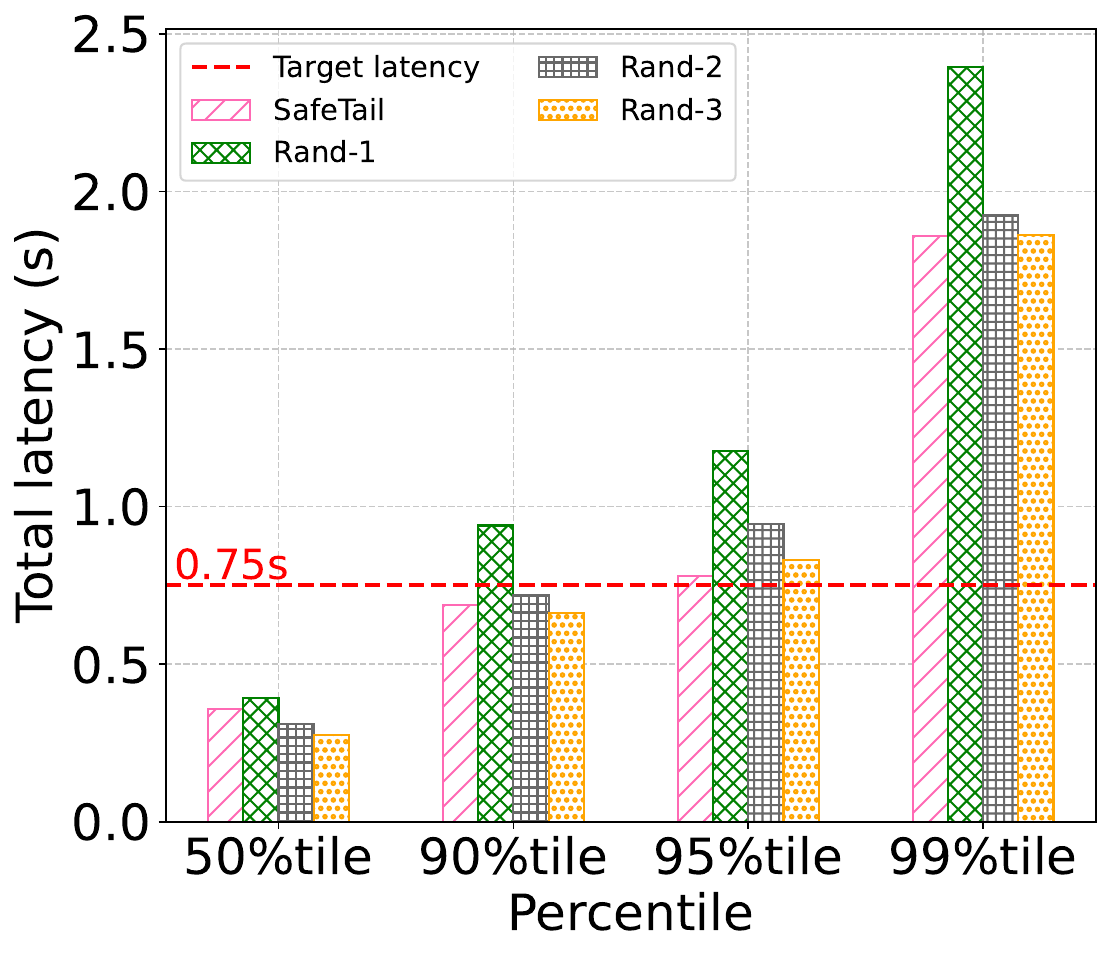}
\includegraphics[width=0.24\textwidth]{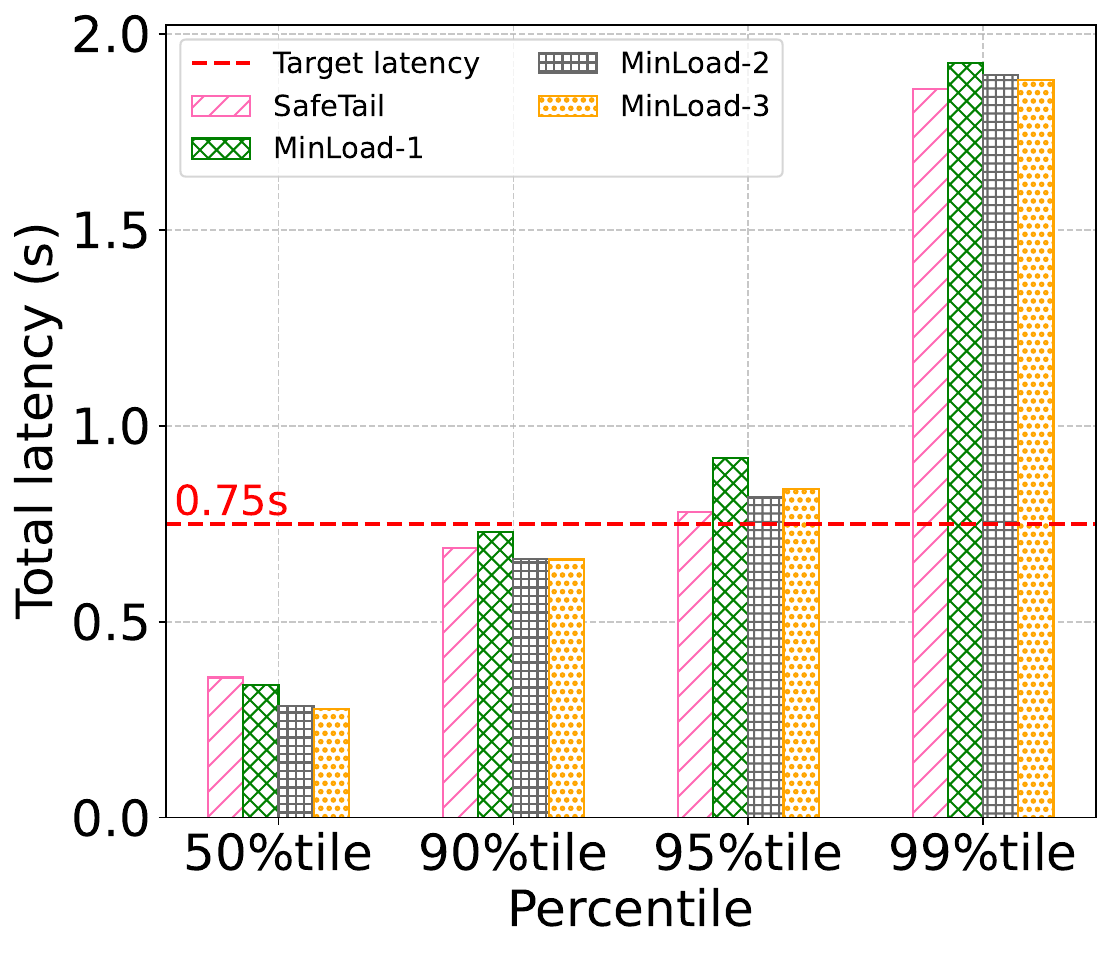}
\includegraphics[width=0.24\textwidth]{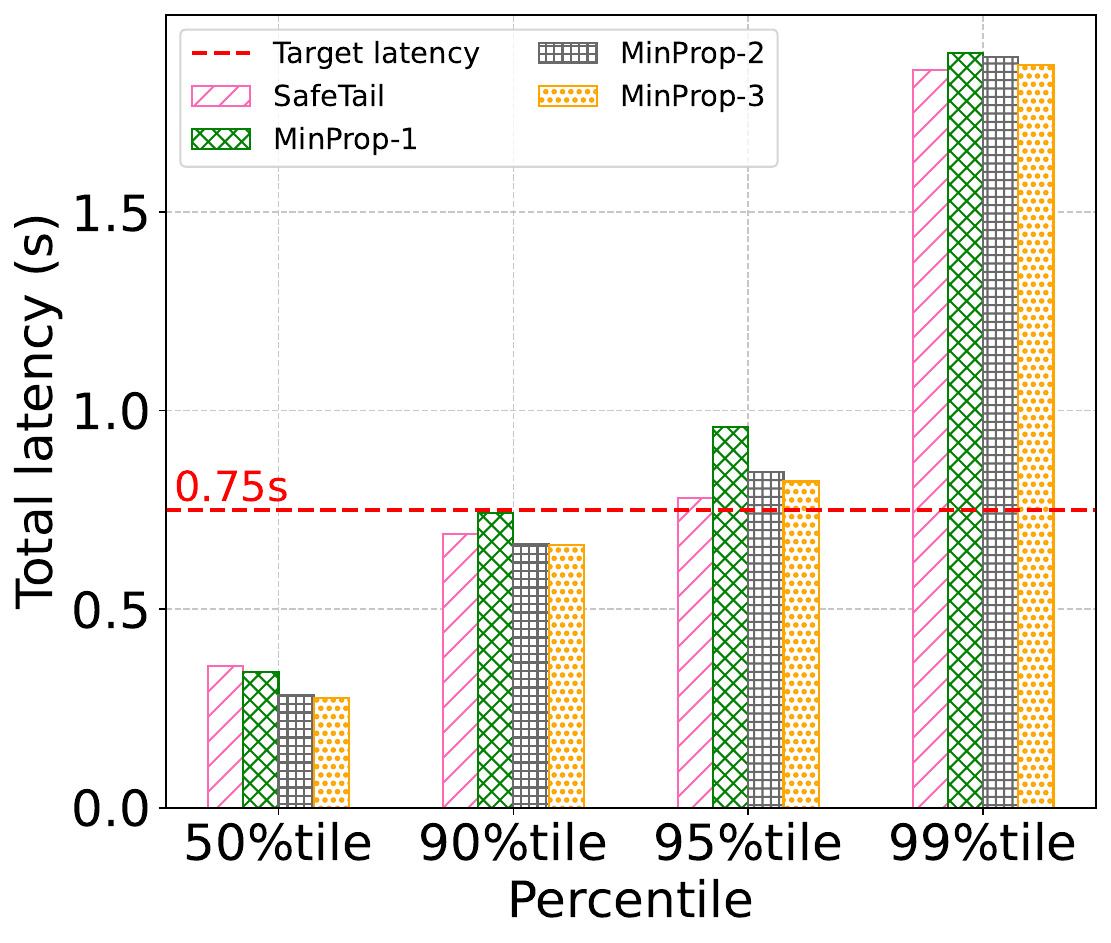}\\
(a)~~~~~~~~~~~~~~~~~~~~~~~~~~~~~~~(b)~~~~~~~~~~~~~~~~~~~~~~~~~~~~~~~~~~~(c)~~~~~~~~~~~~~~~~~~~~~~~~~~~~~~~~~(d)\\~\\
(e) \includegraphics[width=0.24\textwidth]{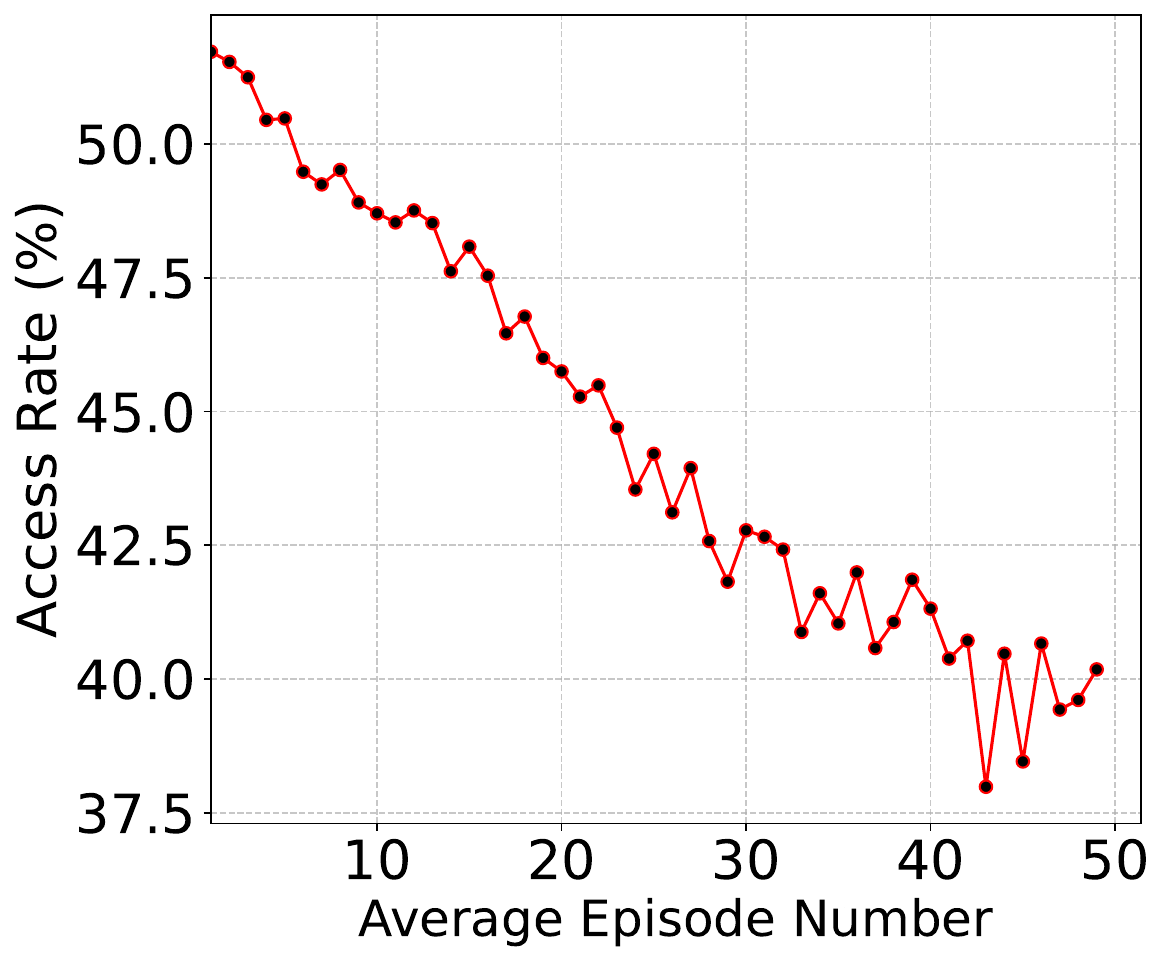} \quad
 (f) \includegraphics[width=0.24\textwidth]{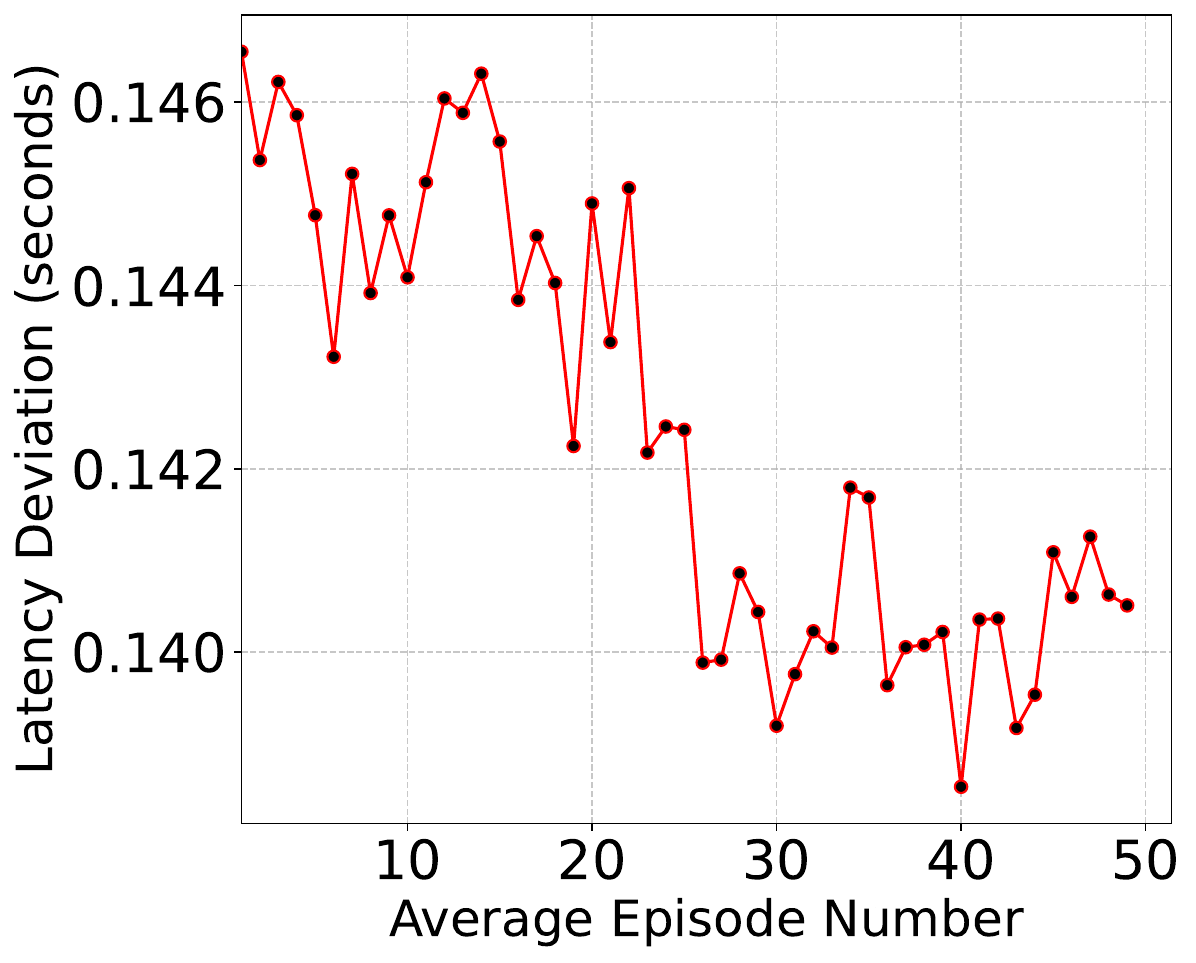}\quad
 (g) \includegraphics[width=0.24\textwidth]{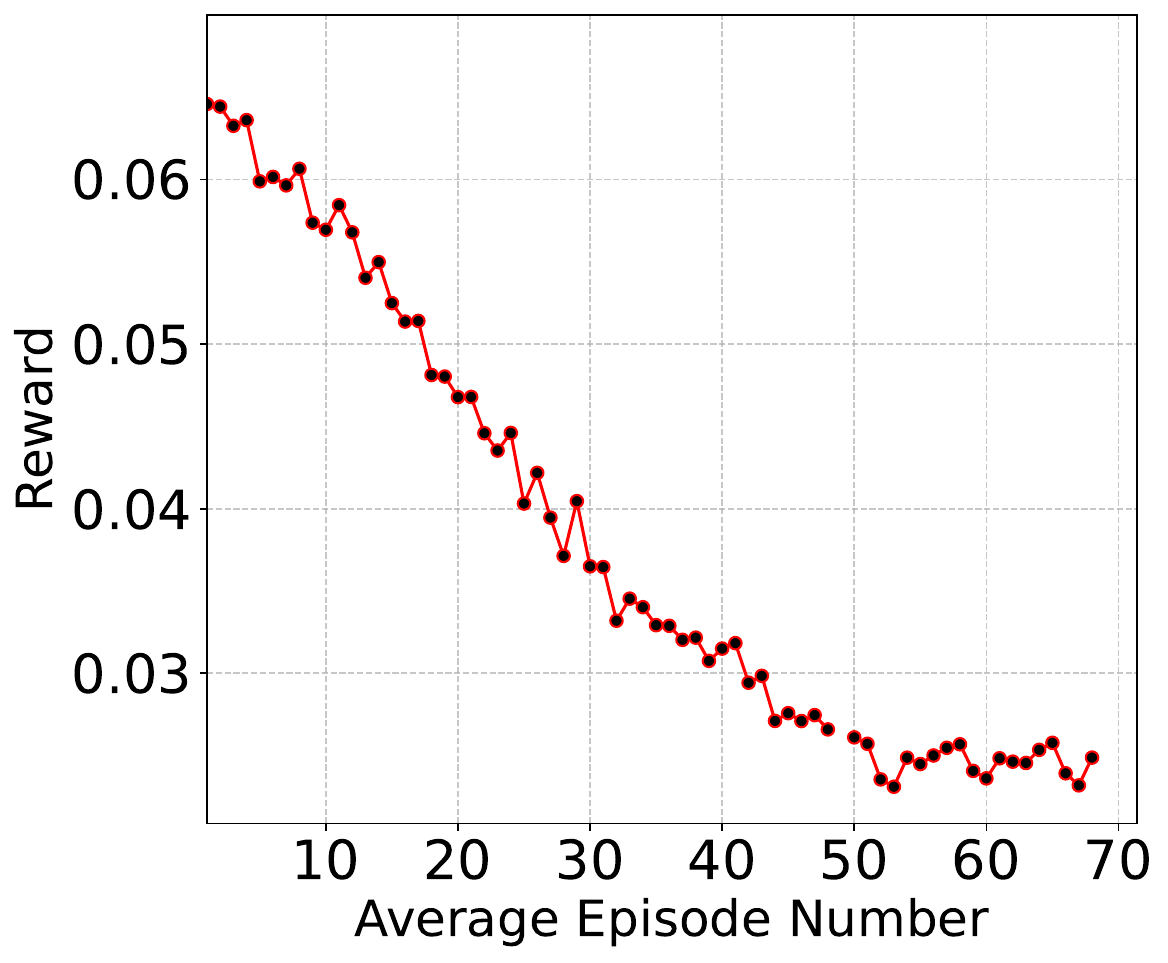}
  \caption{Use Case-3: (a) Latency comparison among SafeTail, target latency and baseline methods, (b) SafeTail versus Rand-x, (c) SafeTail versus MinLoad-x, (d) SafeTail versus MinProp-x, (e) Access rate of SafeTail, (f) Latency Deviation of SafeTail, and (g) Reward distribution of SafeTail.}
     \label{fig:noise_rem_comparison}
 \end{figure*}

\subsubsection{Use Case-2: Instance Segmentation of Images}
\noindent 
A similar trend is observed for this use case regarding median latency, tail latency, variation in access rate, latency deviation, and stabilization of the reward function, with a few exceptions discussed below. 

As shown in Fig. \ref{fig:ins_seg_comparison}(a), while the achieved median latency was lower than the target latency, SafeTail did not always meet the target latency in the case of tail latencies. It is worth noting that even the optimal tail latency exceeded the target latency. This discrepancy arises because the target latency is a heuristic value based on certain assumptions that may not always hold true. The target latency is primarily used for reward computation, but achieving it is not always guaranteed.

Another observation from Fig. \ref{fig:ins_seg_comparison}(b)-(d) is that SafeTail's performance was comparable to Rand-3, MinLoad-3, and MinProp-3. Specifically, for Rand-3, SafeTail showed the worst degradation in median latency, with Rand-3's latency equal to 0.97x that of SafeTail. For MinLoad-3, the most significant degradation was observed in the $90^{\text{th}}$ percentile, where MinLoad-3’s latency was 0.90x that of SafeTail. For MinProp-3, SafeTail's worst performance was seen in the median latency, with MinProp-3 having 0.98x of SafeTail's latency. However, SafeTail outperformed MinProp-3 in other cases. This degradation may be attributed to the baseline methods using a fixed redundancy of three edge servers, leading to higher resource utilization and network congestion. On the other hand, SafeTail selectively uses an average of 2 to 3 edge servers and, in some cases, just 1 edge server. This analysis highlights a clear trade-off between optimizing latency and effectively utilizing resources. 
While the baseline methods achieve lower latency by consistently using multiple edge servers, this approach increases resource utilization and network congestion. In contrast, SafeTail achieves comparable performance with more efficient resource usage by selectively employing fewer edge servers, demonstrating the balance between latency optimization and resource management.

\subsubsection{Use Case-3: Removal of Noise from Audio Files}
An interesting observation can be made from Fig. \ref{fig:noise_rem_comparison}(a)-(d). SafeTail shows a slight degradation compared to the Oracle Optimal value, with Oracle's Optimal being faster by 1.50x, 1.14x, 1.12x, and 1.00x for the median, \pni, \pnf, and \pnn percentiles, respectively. While SafeTail utilized 1-3 edge servers, it predominantly employed only 1-2 edge servers in most cases, as shown in Fig. \ref{fig:noise_rem_comparison}(e). 
It can be observed that for median and \pni percentile latency cases, the baseline methods outperformed SafeTail. However, for the \pnf and \pnn percentile cases, SafeTail performed better. Notably, in the median and \pni percentile cases, even though SafeTail did not surpass the baseline, it achieved latency values better than the target latency. The primary objective of this work is to balance the trade-off between the utilization of a number of edge servers and latency optimization, with a higher priority on reducing tail latency. This result aligned with our objective. In the median and \pni percentile cases, when the achieved latency is better than the target, SafeTail focused on reducing the access rate. Conversely, in the \pnf and \pnn percentile cases, where SafeTail could not meet the target latency, it outperformed the baseline methods even with higher redundancy, because it prioritized latency minimization.

In summary, SafeTail always achieved near-optimal latency while effectively managing resource utilization across all three use cases. In most cases, it outperformed baseline methods without redundancy, and when redundancy was introduced, it delivered competitive median and tail latencies compared to the baseline methods,  while controlling the number of edge servers used. SafeTail's dynamic and selective utilization of edge servers offered a major advantage in resource optimization. Despite minor latency degradations in certain scenarios, SafeTail maintained a stable performance, showcasing a clear balance between latency and resource efficiency.

%% file: 6relatedWork.tex
\section{related work}

\noindent
Existing works that optimize latency focus on three significant contributors to it: studies oriented towards specific latency-sensitive applications, generic scheduling algorithms of edge tasks, and reduction of latency using redundancy. We discuss each of them below:

\noindent \textbf{Optimization of Latency-Sensitive Services on Edge Servers:} Latency-sensitive services are used by applications such as video conferencing, virtual reality and even Industry 4.0. These applications all require optimization of tail latency. A number of other works optimize the latency for critical applications over wireless networks using edge computing \cite{iSapiens,vehicle-slicing,sdn-vehicle}. The work iSapiens \cite{iSapiens} provides a distributed platform to run different smart city applications. The work \cite{vehicle-slicing} schedules tasks for vehicular control on the mobile edge, with suitable slicing of the network resources. The work \cite{sdn-vehicle} proposes utilization of software-defined networking to schedule related to controlling a vehicle. Although these works focus on controlling latency-sensitive applications, unlike SafeTail, they all focus primarily on optimization of median latency and not tail latency.

A few other works also specifically focus on optimizing the computation latency of latency-sensitive services. For example, NeuOS~\cite{neuos} is a system solution designed for running multi-DNN workloads within autonomous systems. COLA optimized the tail latency for Level-4 autonomous vehicles \cite{cola} based on traffic patterns and dynamic latency requirements. They proposed an adaptive data flow and a proactive processing scheme to reduce latency variation in Level-4 autonomous vehicles. TailGuard~\cite{Tail_Guard} ensures that tail latency SLOs are met even under varying workloads by incorporating predictive modeling, real-time monitoring, and adaptive scheduling. Unlike SafeTail, none of them consider the uncertainties present in both the wireless network and the computation time on edge servers. Furthermore, these works also do not utilize redundancy for scheduling services.

\noindent \textbf{Learning based Scheduling for Edge Services:} Learning-based scheduling of edge services is often used to optimize response response times and other quality of service parameters. The survey \cite{survey-edge-scheduling} provides a detailed discussion of the variety of approaches used. For example, the work \cite{uav-mounted-scheduling} uses prioritization of packets for mobile hotspots to schedule tasks offloaded from mobile devices. The work \cite{workload} utilizes deep reinforcement learning to balance the workload across the edge servers. Finally, in the context of edge computing, deep reinforcement learning (DRL) has been employed for caching data in proximity to users~\cite{caching_IoT} and for computation offloading~\cite{DRL_computation}, ~\cite{collaborative_MEC}. Notably,~\cite{DRL_computation} utilizes DRL to identify the components to offload, departing from traditional optimization methods. This approach assumes a linear relationship when mathematically modeling the edge compute system. These  works focus on caching data, data offloading, and optimizing latency. However, different from SafeTail, they do not focus on scheduling of services to reduce tail latency. 


\noindent\textbf{Utilization of Redundancy to Reduce Latency:} The concept of redundancy for enhancing reliability is widely employed in data centers~\cite{duplication-conext,tail-nsdi}. For example, the work \cite{stragglers} performs root cause analysis of high tail latency in longer jobs and identify that the primary culprit behind such latency issues is high server utilization. This insight enables the prediction of potential slower jobs early into their computation, enabling efficient management and optimization. In \cite{straggler-replication}, an algorithm is proposed to estimate the latency and cost tradeoff based on the empirical distribution of task computation time. This shows that a modest level of task replication can diminish both latency and the expenses associated with computing resources. Furthermore, \cite{less-provisioning} balances the requirement of tail latency with that of reduction of redundancy, similar to ours. However, they depend on integer programming and online algorithms, instead of using a learning-driven approach. These works also focus on cloud computing, and not on edge computing, where network latencies are higher and more uncertain.

In the context of edge computing, duplication ~\cite{chang2018adaptive} is used on edge servers to shorten the transmission time and improve the quality of service (QoS). Duplicate aware scheduling was proposed to duplicate every task. The work ~\cite{Replication_Proactive} proactively senses the network and compute resources available to ascertain the need for duplication before scheduling services on the edge. In~\cite{Replication_MEC}, authors explored the use of computation duplication to edge servers to accelerate the download of results. The work \cite{Replication_Vehicular} focuses on reducing latency and reliability in edge computing with inherent system uncertainty. A key difference of these techniques from our approach is that they do not have a specific \textit{target} latency. Our notion of target latency ensures that the level of duplication remains limited and thus, reduces the amount of resource utilization.

%% file: 7conclusion.tex
\section{Limitation and Future Work}
\noindent
SafeTail currently has the following limitations. First, it has been evaluated on a homogeneous set of edge servers, where all servers have identical computing and network resources. While this is common in prior studies, some research has shown that resource heterogeneity in edge servers can help reduce latencies. Although we considered homogeneous edge servers in our simulation, our framework captures the dynamic state of each server, which varies across the edge servers. Therefore, we believe our approach can be seamlessly extended to heterogeneous settings as well. In future work, we intend to expand our experiments to consider heterogeneous environments. Additionally, our current framework assumes homogeneous services for execution, a constraint we also plan to relax in future work.

Second, our framework is currently user-centric, with SafeTail operating on the user side and making redundancy decisions based solely on the user's needs. However, this approach may occasionally lead to increased overall resource consumption, although we mitigate this by reducing redundancy when possible. In the future, we aim to extend our work to optimize tail latency across services by considering the needs of all users in the network.

Third, we did not model the waiting time for each server. We assume that if a server cannot accept a request, it simply discards it. Addressing this limitation will also be a focus of our future work.


\section{conclusion}  
\noindent
In this paper, we introduced SafeTail, a reward-based deep learning framework designed to tackle the challenge of reducing tail latency in latency-sensitive services. SafeTail aims to optimize service execution latency through adaptive redundancy, intelligently managing the use of additional edge servers to achieve this goal. The framework dynamically adapts to the changing conditions of edge servers and service demands, deploying redundancy only when necessary to minimize tail latency while avoiding excessive resource use and network congestion.

Our experimental results revealed that SafeTail significantly improved both median and tail latency compared to existing baseline methods. We performed extensive trace-driven simulations across diverse applications, including object detection, image segmentation, and audio noise removal. These simulations demonstrated that SafeTail effectively reduced service latency, particularly tail latency, while skillfully balancing latency and resource utilization.